\documentclass{frontiersSCNS} %
\usepackage{url,hyperref,microtype,subcaption,color}
\usepackage[onehalfspacing]{setspace}
\def\keyFont{\fontsize{8}{11}\helveticabold } \def\firstAuthorLast{Chamanbaz
  {et~al.}}
\def\Authors{Mohammadreza Chamanbaz\,$^{1,*}$, David Mateo\,$^{1}$, Brandon
  M. Zoss\,$^{2}$, Grgur Toki\'c\,$^{2}$, Erik Wilhelm\,$^{1}$, Roland
  Bouffanais\,$^{1,2,*}$, and Dick K. P. Yue\,$^{2}$}
\def\Address{$^{1}$Applied Complexity Group, Singapore University of Technology and Design, Singapore, Singapore\\
  $^{2}$Department of Mechanical Engineering, Massachusetts Institute of
  Technology, Cambridge, MA, USA}
\def\corrAuthor{Roland Bouffanais} \def\corrAddress{Applied Complexity Group,
  Singapore University of Technology and Design, Singapore 487372, Singapore}
\def\corrEmail{bouffanais@sutd.edu.sg}
\begin{document}
\onecolumn \firstpage{1}
\title[Swarm-Enabling Technology for Multi-Robot Systems]{Swarm-Enabling
  Technology for Multi-Robot Systems}
\author[\firstAuthorLast
]{\Authors} 
\address{} 
\correspondance{} 
\extraAuth{}
\maketitle
\begin{abstract}
  Swarm robotics has experienced a rapid expansion in recent years, primarily
  fueled by specialized multi-robot systems developed to achieve dedicated
  collective actions. These specialized platforms are in general designed with
  swarming considerations at the front and center. Key hardware and software
  elements required for swarming are often deeply embedded and integrated with
  the particular system. However, given the noticeable increase in the number
  of low-cost mobile robots readily available, practitioners and hobbyists may
  start considering to assemble full-fledged swarms by minimally retrofitting
  such mobile platforms with a swarm-enabling technology.  Here, we report one
  possible embodiment of such a technology---an integrated combination of
  hardware and software---designed to enable the assembly and the study of
  swarming in a range of general-purpose robotic systems. This is achieved by
  combining a modular and transferable software toolbox with a hardware suite
  composed of a collection of low-cost and off-the-shelf components. The
  developed technology can be ported to a relatively vast range of robotic
  platforms---such as land and surface vehicles---with minimal changes and
  high levels of scalability. This swarm-enabling technology has successfully
  been implemented on two distinct distributed multi-robot systems, a swarm of
  mobile marine buoys and a team of commercial terrestrial robots. We have
  tested the effectiveness of both of these distributed robotic systems in
  performing collective exploration and search scenarios, as well as other
  classical cooperative behaviors.  Experimental results on different swarm
  behaviors are reported for the two platforms in uncontrolled environments
  and without any supporting infrastructure.  The design of the associated
  software library allows for a seamless switch to other cooperative
  behaviors---e.g. leader-follower heading consensus and collision avoidance,
  and also offers the possibility to simulate newly designed collective
  behaviors prior to their implementation onto the platforms. This feature
  greatly facilitates behavior-based design, i.e., the design of new swarming
  behaviors, with the possibility to simulate them prior to physically test
  them.
  \tiny \keyFont{ \section{Keywords:} Swarm Robotics, Distributed Multi-Robot
    Systems, Cooperative Control, Flocking, Distributed
    Communication} 
\end{abstract}

\section{Introduction}
A swarm robotics system consists of \emph{autonomous robots} with \emph{local
  sensing and communication capabilities}, lacking \emph{centralized control}
or access to \emph{global information}, situated in a \emph{possibly unknown
  environment} performing a collective
action~\citep{brambilla2013swarm}. Based on this definition, one can easily
distinguish swarm robotics systems from other multi-robot
approaches~\citep{Iocchi2001}. It is not uncommon for multi-robot systems to
lack some form of decentralization at the computation, communication and/or
operation levels~\citep{vicsek_platform}. Swarm robotics systems are often
inspired by natural systems in which large numbers of simple agents perform
complex collective behaviors---e.g.  schooling fish, flocking birds---through
repeated local interactions between themselves and their
environment~\citep{bouffanais_design_2016}. The swarm robotics design paradigm
allows a multi-robot system to overcome the aforementioned limitations (local
sensing, lack of centralized control, etc.) and operate autonomously in a
coordinated fashion.

One main challenge in artificial swarming is the design of systems that, while
maintaining decentralized control, have agents capable of {\it (i)} acquiring
local information through sensing, {\it (ii)} communicating with at least some
subset of agents, and {\it (iii)} making decisions based on the
dynamically-gathered sensed data.  While decentralization denies the agents
the benefits of a large central computation and/or communication hub, it
affords the system robustness.  The system is thus capable of performing
global collective actions under a wide range of group sizes (scalability),
despite the possible sudden loss of multiple agents (robustness), and under
unknown and dynamic circumstances (flexibility)~\citep{brambilla2013swarm}.

There have been a number of notable efforts to design robotic swarms.  The
kilobot~\citep{Kilobot}, e-puck~\citep{puck}, Swarm-bots~\citep{swarm_bots},
marXbot~\citep{marXbot}, Alice~\citep{Alice}, iAnt~\citep{iant},
Scarab~\citep{scarab}, I-Swarm~\citep{iswarm}, r-one~\citep{r_one} and Pi
Swarm~\citep{pi_swarm} are only a few examples of different swarm/distributed
multi-robot platforms that have been developed. Each platform explores the
potential and feasibility of a few aspects of swarming and not all of them
fulfill all requirements of a robotic swarm system---according to the
definition above---or are capable of operating in real and uncontrolled
environments without any supporting infrastructure.  The technological
advances implemented on each platform make it challenging to port to others
due to the specificities of each robot. Indeed, these platforms have been
conceived with swarming considerations at the core of their design. As a
consequence, the central components necessary to enable swarming are fully
integrated inside the robots. Moreover, the software layer is often highly
dependent on the hardware specifications owing to the original co-design
process of both hardware and software. Although this deep technical
integration lacking modularity provides platforms that can easily swarm, it
also is a serious impediment to its portability to other mobile platforms.

Practitioners with access to multiple units of an existing autonomous robotic
platform---including commercial ones---and seeking to study swarming have
currently no other option than devising their own custom-made framework. A
common alternative used by a number of research groups consists in simply
devising swarming experiments that can be achieved with existing commercial
swarm robotics platform (e.g. kilobot, e-puck, etc.). Although this
alternative has the advantage of piggybacking on well-tested platforms,
thereby saving a significant amount of time, it nonetheless limits the ability
to design very original experiments. Our proposed technological solution aims
at filling this gap by offering a platform-agnostic framework---an integrated
combination of a hardware suite with a software layer---easily portable to a
host of hardware platforms and facilitating the implementation of various
swarm algorithms, while reducing the burden associated with platform-dependent
interfacing.

Here, we report a unified platform-agnostic hardware/software tool capable of:
(\textit{i}) assembling and transforming a collection of basic mobile robots
into full-fledged swarms, and (\textit{ii}) achieving versatile swarming
behaviors using an easily programmable software library. The modular nature of
both the hardware suite and software layer allows for possible evolution,
upgrade and extension of the technology independently from the specifications
of the mobile platform. We present some preliminary results validating the
technology on two vastly different swarming systems. This swarm-enabling
technology is a combination of hardware/software that allows a wide range of
multi-robot platforms to perform versatile and responsive swarming. This is
achieved by providing each agent with an additional interface hardware built
from low-cost and off-the-shelf components which is integrated with a
specifically-developed general purpose software library. We believe that this
technology that decouples swarming considerations from robotics ones could be
of interest to both researchers and educators primarily interested in the
study of artificial swarming.

The software---written in Python but also tested in C++---is designed in a
modular way so as to make the technology portable between platforms as
seamlessly as possible, i.e. with minimal hardware/software changes. To assess
the effectiveness of the proposed technology, we used it on two different
platforms operating in uncontrolled environments: a differential drive robot
and a water surface platform (sensing mobile buoy). Furthermore, a number of
collective behaviors such as heading consensus, perimeter defense and
collective marching are tested on these platforms, thereby confirming the
versatility of our swarm-enabling technology.

\section{Material \& Methods}

%
To enable swarming, it is essential to achieve distributed communication and
decentralized
decision-making~\citep{vigelius2014multiscale,hamann2014derivation,%
  valentini2014self,%
  valentini2015self,%
  valentini17:_best_probl_robot_swarm}.  For instance, natural swarms achieve
self-organizing behaviors and decentralized decision-making by means of
distributed information exchanges through local
signaling~\citep{camazine03:_self_organ_in_biolog_system} associated with
sometimes sophisticated signaling mechanisms and trophic
interactions~\citep{dusenbery92:_sensor_ecolog}.  Note that signaling refers
to communication involving sensory capabilities. Swarm robotics systems mimic
natural swarms in that communications between units is restricted to local
information exchanges through short-range
interactions~\citep{brambilla2013swarm}. More generally, locality of
communication in space and time between individual platforms leads to
distributed communication. Therefore, platforms should be able to establish a
dynamic and possibly switching communication networks and process information
locally, using solely the computation capabilities on-board each individual
agent.

\subsection{Computational Unit}

%
Most multi-robot platforms have limited computational resources only to read
sensors data from the robot and send them to a remote central unit. The remote
processor makes appropriate decisions based on its supervisory control
algorithm and sends commands back to the robot. Since we need the technology
to be easily portable to many different robotic platforms, we use a dedicated
processing unit independent from the robot's sensing and actuation processor
as the ``brain'' of each agent.  In our setup, computations are performed by
Raspberry Pi/BeagleBone, single-board computers (SBCs) equipped with a number
of multi-purpose inputs and outputs. These two SBCs provide ample
computational resources for their size given the tasks at hand. Note that one
can also consider using a different SBC such as
Gumstix\footnote{\url{https://www.gumstix.com/}} to serve the same
purpose. The recently released Raspberry Pi ``0'' (resp. ``3'') offers a very
low-cost solution at half-credit card size (resp. high computational power and
increased interfacing possibilities). If even more computational power is
required, one may consider the Odroid SBC or ultimately the GIGABYTE.

This computational unit receives all sensor data from the robot and neighbors'
information through the communication network in order to make appropriate
decision based on the swarm algorithm, which is embedded in the software
toolkit detailed below; this latter part is pre-loaded onto each platform
prior to any collective operation.

\subsection{Distributed Communication}
%
From a theoretical viewpoint, distributed communication can be better analyzed
and understood using network theoretical concepts---even in the absence of a
real physical communication network as in the case of flocking birds and
schooling fish~\citep{bouffanais_design_2016}. Recently, studies of such
signaling networks in swarms have revealed the need for specific structural
properties of the network---in terms of degree distribution, shortest path and
clustering coefficient---in order to achieve effective consensus-reaching
dynamics~\citep{sekunda16:_inter,shang14:_influen} and high dynamic
controllability of the swarm by a given subset of driving
agents~\citep{komareji13:_resil_contr_dynam_collec_behav}.

From a practical viewpoint, platforms should be able to establish a
dynamic---i.e., switching---communication network. Recently, studies of such
temporal networks~\citep{holme12:_tempor} in swarms have revealed the need for
specific structural properties of the network---in terms of degree
distribution, shortest path and clustering coefficient---in order to achieve
effective consensus-reaching dynamics
\citep{shang14:_influen,sekunda16:_inter} and high dynamic control of the
swarm by a given subset of driving agents
\citep{komareji13:_resil_contr_dynam_collec_behav}.  With a mesh network, all
agents are identical (from the network viewpoint) and can exchange information
with a specific set of neighbors (e.g., metric, topological, Voronoi
neighborhoods) directly without involving a third agent or going through a
central hub or router.  This grants the system robustness, as the loss of a
subset of agents does not have a critical impact on the operation of the rest,
assuming the nodes are within the limited communication range.  This stands in
stark contrast with the star network configuration, where the loss of the
routing agent will halt the operation of the whole system. The same scenario
would hold if the agents communicate in a mesh network but rely on a
centralized computational hub to process information as is actually the case
with many multi-robot systems, see~\citep{vicsek_platform} for further
discussion.

\begin{figure}[htbp]
  \centerline{\includegraphics[width=1\textwidth]{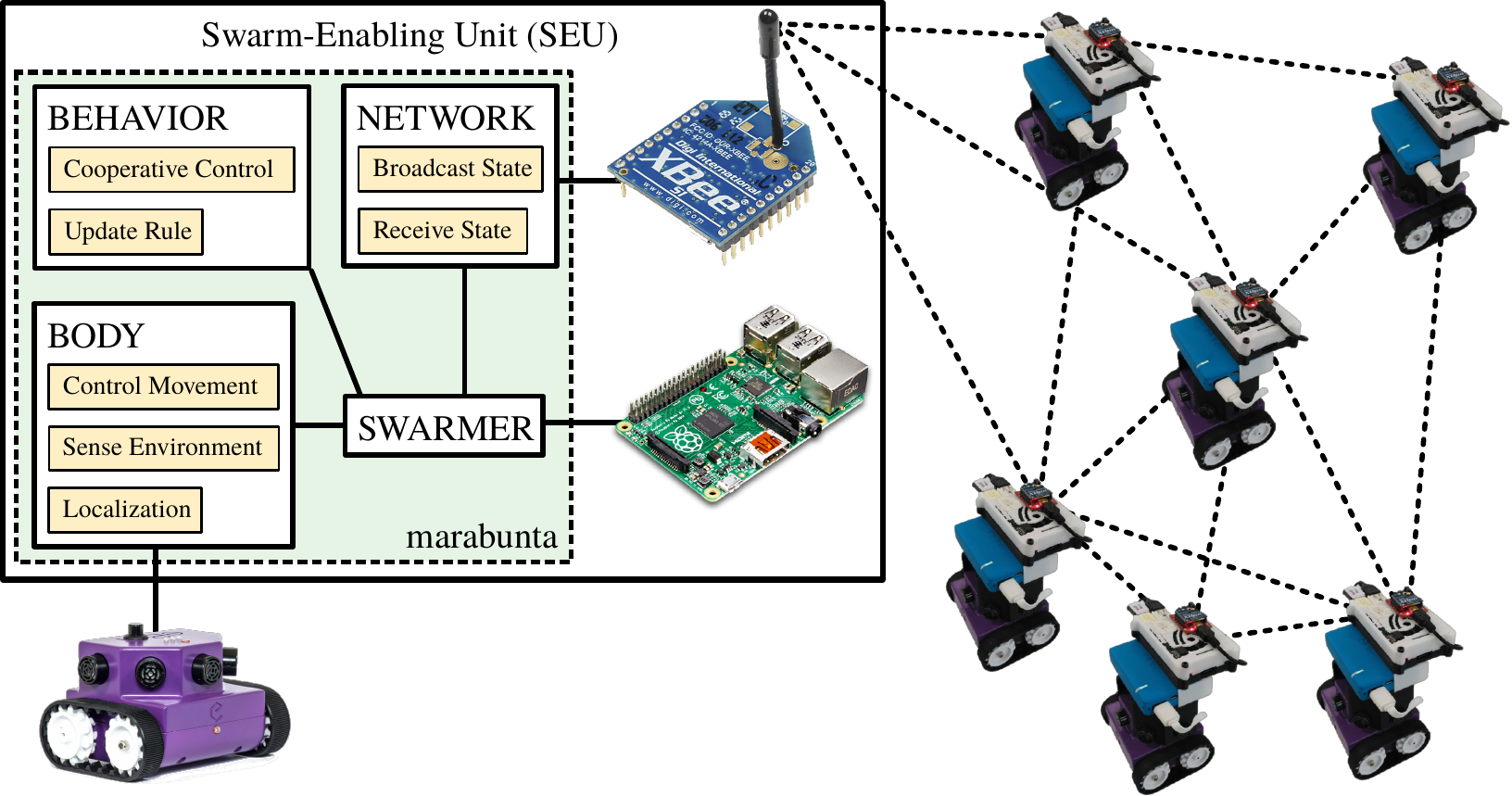}}
  \caption{Block diagram of the Swarm-Enabling Unit (SEU).  The SEU serves as
    a bridge between a particular robot and the swarm, and is composed by a
    communication module and a processing unit running code using the
    marabunta module.  At the software level, each agent (or swarmer) is
    composed by three elements. First, the ``body'' interacts with the robot
    to control its movement and gather information from its state and sensed
    environmental data.  Second, the ``network'' interacts with the
    communication module to broadcast the current state of the agent to the
    swarm and to gather information from other agents' state.  Third, the
    ``behavior'' contains the cooperative control strategy.  In its most
    elementary form, the behavior is implemented by an update rule that
    defines the movement of the robot for a certain time window, given the
    current state of the robot and the data gathered from the body and the
    network.  }
  \label{fig:block diagram of the system}
\end{figure}

The crucial component required to achieve a fully decentralized system is the
communication device. We use XBee modules~\citep{XB} to establish a mesh
network.  The communication network is based on a metric interaction in which
information is being communicated between agents within the communication
range of one another (typically 300 m line of sight).  The device is
configured in broadcast mode in which the information sent by each agent is
received by all neighbors within the communication range.  The range of
communication depends on various factors such as the output power of the
module and the type of obstacles blocking the radio frequency wave. The
hardware block diagram of this swarm-enabling unit (SEU) is shown in
Fig. \ref{fig:block diagram of the system}.

A natural concern with such a dynamic and distributed communication network is
its effectiveness in maintaining a sustained flow of information among
swarming agents. This was analyzed and quantified during swarming experiments
with the BoB system with buoys continuously broadcasting their state at 0.1
Hz.  The expected communication range is about $310$~m and the modules are
capable of relaying messages through multiple hops in the network, which means
that in principle any module can broadcast their state globally to all the
agents in the collective.  However, our experiments show that when tens of
buoys are operating, the communication is far from perfect, and the effective
communication range is significantly smaller. The ratio of successful
communications obtained during a typical field experiment is presented in
Table~\ref{tab}. These results provide a measure of the effective
communication range in a large and dynamic network of mobile XBee units. As
one increases the number of buoys deployed, interferences between them will
cause more messages to drop. This presents a clear example of why the
distributed control algorithm should be designed to provide a robust
collective behavior against imperfect communication.
\begin{table}[h]
  \textbf{\refstepcounter{table}\label{tab}Table \arabic{table}.}{ Relative successful communications established
    between two buoys separated by a certain distance inside a dynamic collective of
    $N$ buoys spread with a mean nearest-neighbor distance of $\langle r_0
    \rangle$.}
  \processtable{ }
  {\begin{tabular}{lll}\toprule
      $(N,\langle r_0\rangle (\text{in m}))$ & Distance (m) & Communication success\\\midrule
      (20,19.5) & 10 &  96~\%\\
      (20,19.5) & 40 & 91~\%\\
      (20,19.5) & 80 & 88~\%\\
      (20,19.5) & 120 & 84~\%\\
      (40,6.9) & 10 &  89~\%\\
      (40,6.9) & 40 &  86~\%\\
      (40,6.9) & 80 &  81~\%\\
      (40,6.9) & 160 &  66~\%\\\botrule
    \end{tabular}}{}
\end{table}

\subsection{Cooperative Control Strategy} \label{sec:swarm algorithms}
%
The achievement of effective collective behaviors by a multi-robot system
requires fully decentralized control algorithms. Such cooperative control
strategies have received particular attention from different scientific
communities with different aims and goals: first by the computer graphics
community~\citep{reynolds}, then followed by the physics
community~\citep{viscek2012}. The control community subsequently established a
formal
framework~\citep{olfati-saber07:_consen_cooper_networ_multi_agent_system,renbook,jad},
which has been put into practice and expanded by multi-robot systems and swarm
robotics community~\citep{brambilla2013swarm,turgut2008self}. Recently, Vicsek
and his collaborators have established the connection between dynamical update
rules of locally interacting agents and cooperative control strategies for
flocks of autonomous flying robots~\citep{viragh2014flocking}. This endeavor
was fueled by an intense research activity from biologists and physicists who
have sought to identify local update rules at the agent level, which result
into observed collective animal
behavior~\citep{bouffanais_design_2016,viscek2012}. This acquired knowledge
enabled novel biologically inspired approaches to the design of cooperative
control strategies. Such an approach has been successfully implemented and
tested on a small flock of 10 quadcopters~\citep{vicsek_platform}, using GPS
for localization, and the same distributed communication paradigm as the one
reported here. Using the taxonomy introduced by Brambilla~\textit{et
  al.}~\citep{brambilla2013swarm}, our decentralized cooperative control
strategy follows a behavior-based design approach in a similar vein as
Vicsek~\textit{et al.}~\citep{viragh2014flocking}.

In what follows, we discuss several swarm algorithms coded and tested on
different platforms.

\subsubsection{Consensus}\label{sec:heading consensus}
In a consensus algorithm, the participating agents seek to have their state
variable---in the present case their heading---converging towards a common
value; this latter value is not known a priori and is entirely the outcome of
this self-organizing process~\citep{consensus_survey}.

In the particular case of collective motion, agents can aim at aligning their
direction of travel, leading a well-aligned flock of agents all traveling in
the same direction. Another common example is aggregation. As its name
implies, this dynamical behavioral rule leads the agents to collectively
undergo an aggregation process. Such collective aggregation is extremely
common and important in natural swarms (e.g., insects and microorganisms such
as amoebae~\citep{bouffanais10:_hydrod_of_cell_cell_mechan}) and is also very
useful with distributed multi-robot systems during certain phases of
deployment in the field.

Denote the set of agents whose information is available to agent $i$ as
$\mathcal{N}_i$, a consensus protocol can be summarized as
\begin{equation}\label{eq: consensus original}
  x_i[k+1] = \sum_{j\in\mathcal{N}_i\cup\{i\}}\alpha_{ij}[k]x_j[k],
\end{equation}
where $x_i[k]\in\mathbb{R}$ is the state of agent $i$ at time $k$ and
$\alpha_{ij}[k]>0$ is a desired weighting factor. Considering the heading
$\hat{\theta}_i=(\cos\theta_i,\sin\theta_i)$ and choosing $\alpha_{ij}=
\frac{1}{N_i+1}$ where $N_i=|\mathcal{N}_i|$, (\ref{eq: consensus original})
reads
\begin{equation}\label{eq:theta consensus}
  \hat{\theta}_i[k+1] = \sum_{j\in\mathcal{N}_i[k]\cup\{i\}}\frac{1}{N_i+1}\hat{\theta}_j[k].
\end{equation}
Essentially, the target heading at $k+1$ is defined as the average heading of
the agent itself and its neighbors at time $k$.  This protocol only involves
local information exchange and is guaranteed to generate a global common
consensus if the swarm is connected~\citep{jadbabaie2003coordination}.

\subsubsection{Perimeter Defense } \label{sec:perimeter defense}

The perimeter defense, or all round defense algorithm, for swarming agents
consists in having them self-organize so as to maximize the perimeter covered
in an unknown dynamic two-dimensional environment. To obtain this result over
local information transfers, the agents maximize the distance between
themselves and their neighbors, giving a larger weight to the closer agents.
This can be encoded as the dynamical rule
\begin{equation}\label{eq:perimeter defense}
  p_i[k+1] = \sum_{j\in\mathcal{N}_i}\frac{p_j[k]-p_i[k]}{|p_j[k]-p_i[k]|^2},
\end{equation}
where $p_i[k]\doteq(x_i[k],y_i[k])$ is the position of agent $i$ in Cartesian
coordinates at time $k$. In Eq.~\eqref{eq:perimeter defense}, the target
heading is obtained from $p_i[k+1]$ and it is designed such that a larger
weight is given the closer the neighbor is. Note that Eqs.~\eqref{eq:theta
  consensus} and~\eqref{eq:perimeter defense} are purely Markovian, involving
only local information in both space and time without any assumption of
information about prior or future states. Thus, any such algorithm comes with
implicit flexibility, as the swarm can operate under dynamic environments.

\subsubsection{Environment Exploration}

For the purposes of environment exploration, we combine heading consensus and
a modified attraction-repulsion behavior with individually manifested heading
goals. Unlike the perimeter defense, the goal here is to provide a dynamic
two-dimensional spatial coverage for sensing and environment reconstruction
and prediction purposes.  Each member has the autonomy to diverge from
collective behaviors in order to investigate their surroundings, yet maintain
underlying neighbor-to-neighbor interactions.

\begin{figure}[htbp]
  \centering
  \begin{subfigure}{0.32\columnwidth}
    \centering
    \includegraphics[width=1.0\columnwidth]{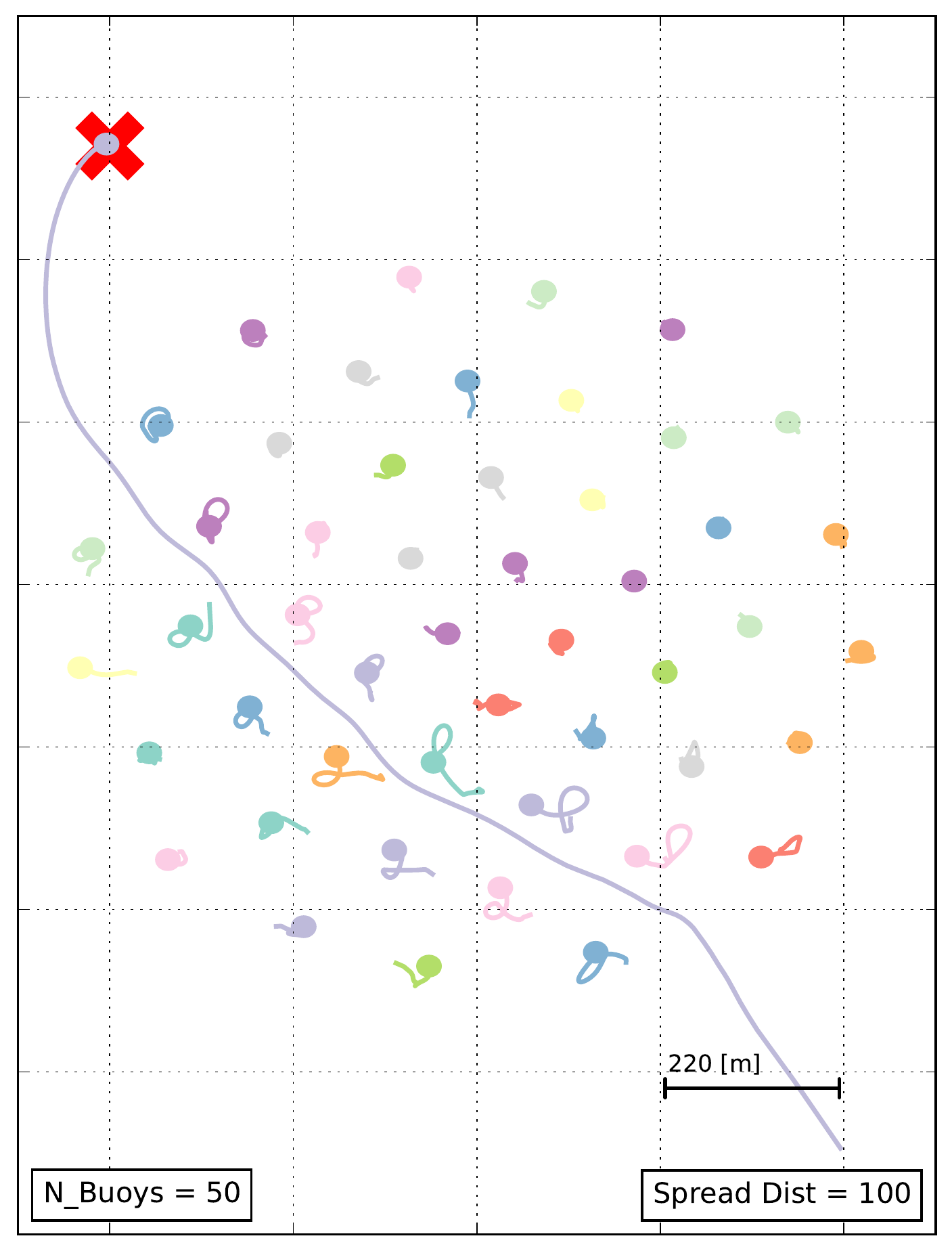}
  \end{subfigure}
  ~
  \begin{subfigure}{0.32\columnwidth}
    \centering
    \includegraphics[width=1.0\columnwidth]{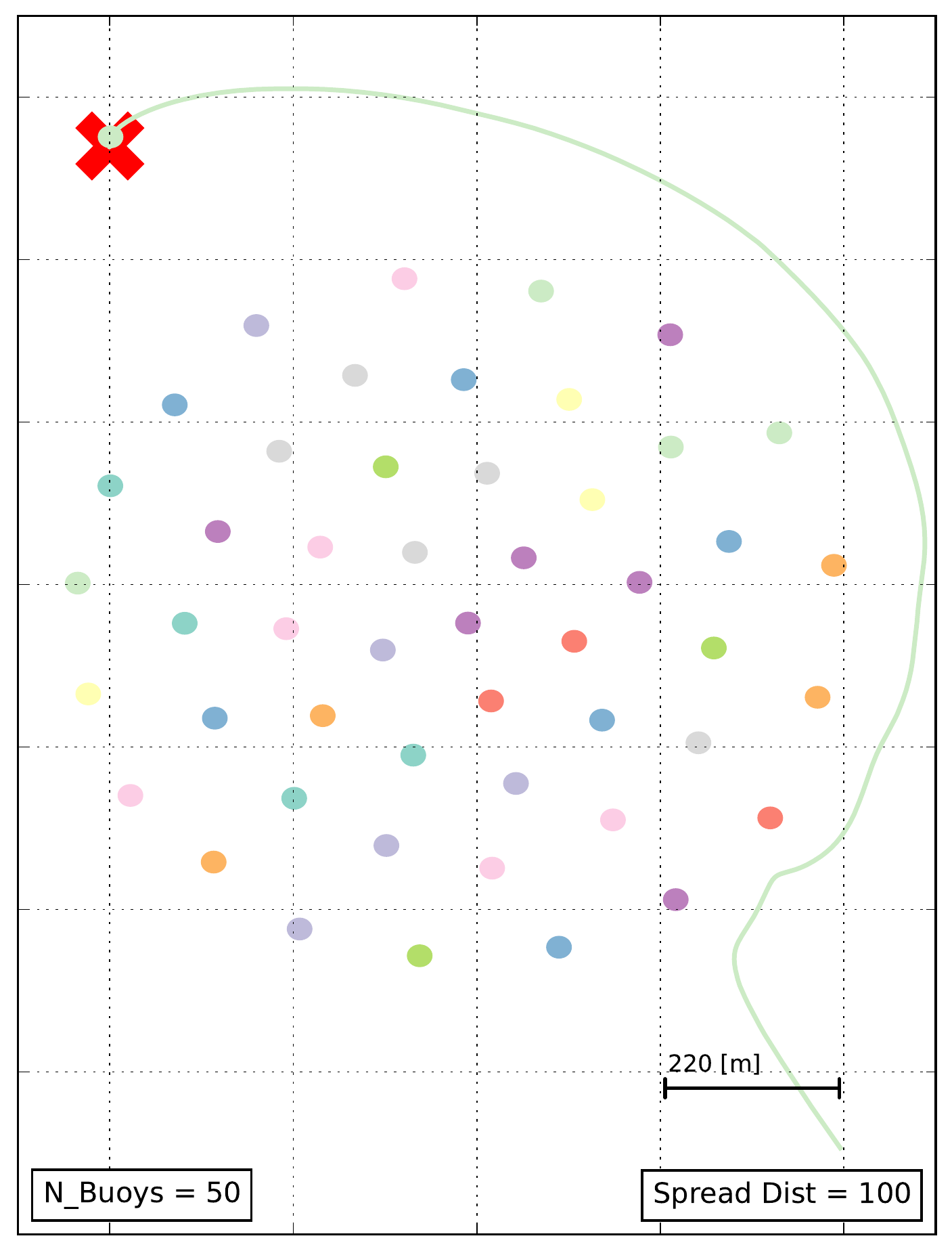}
  \end{subfigure}
  ~
  \begin{subfigure}{0.32\columnwidth}
    \centering
    \includegraphics[width=1.0\columnwidth]{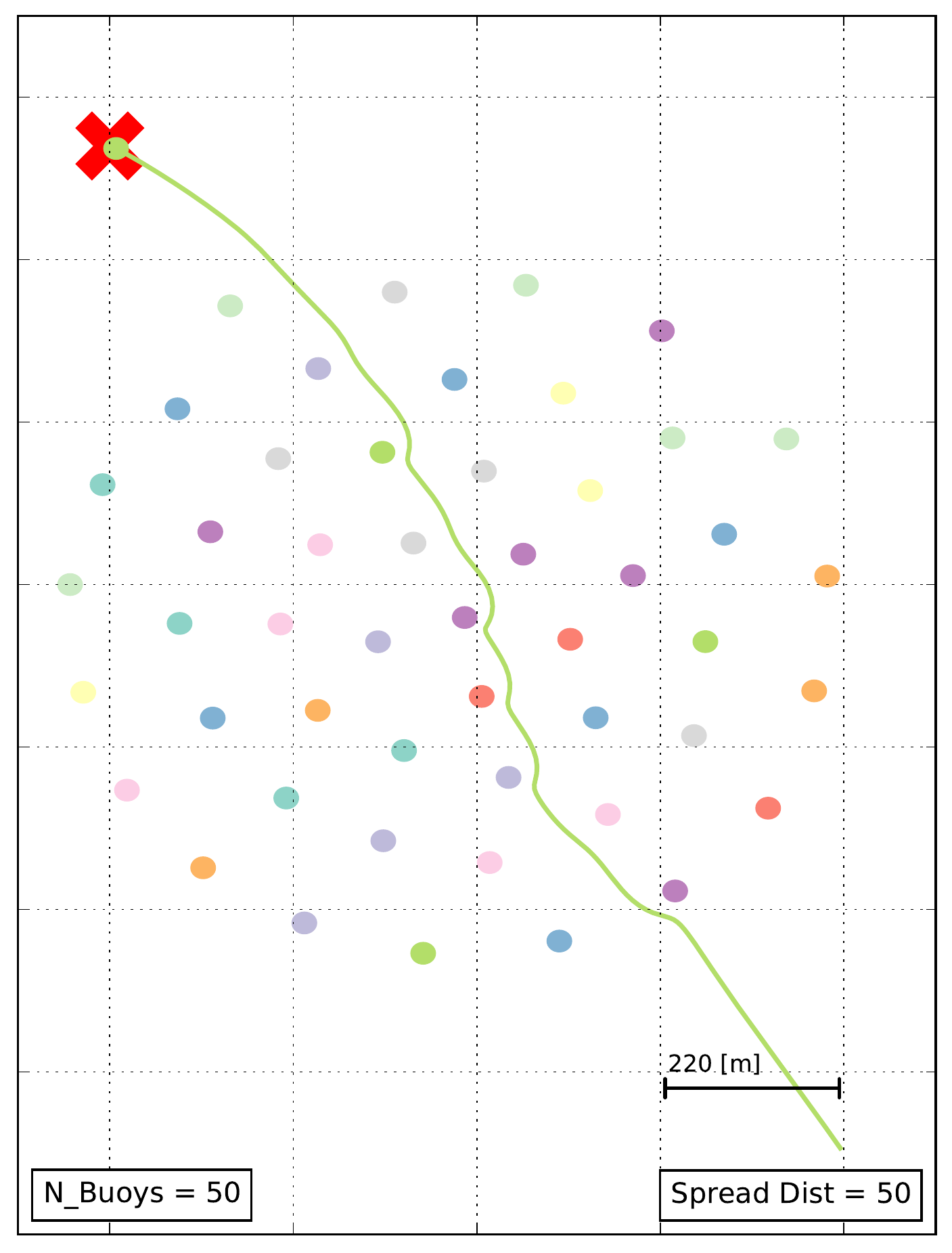}
  \end{subfigure}
  \caption{Examples of simulated avoidance behaviors based on
    Eq.~\eqref{eq:BoBSwarm}, with $H_i = 1$, $H_j = 0$,$(p_0)_j = 100$~m, $j
    \neq i$. The mission of agent $i$ is to reach the goal (red cross) when
    the path is blocked by other agents in the swarm (starting from a position
    at the lower right corner). Left: Yielding behavior. The equilibrium
    distance $(p_0)_i = 100$~m of agent $i$ is too large for it to go through
    the swarm without reconfiguration. In response, the agents in the swarm
    yield as agent $i$ moves towards the goal through the swarm, only to
    circle back to their equilibrium positions based on
    \eqref{eq:BoBSwarm}. Center: Swarm fixed in place ($v_j\equiv 0$). The
    equilibrium distance $(p_0)_i = 100$~m is too large for the agent $i$ to
    go through the swarm, so it goes around. Right: Swarm fixed in place
    ($v_j\equiv 0$). The equilibrium distance $(p_0)_i = 50$~m is small enough
    for the agent $i$ to go through the swarm.}
  \label{fig:BoBavoidance}
\end{figure}

Self-assembling within a group relies on three basic aspects: equilibrium
distance between neighbors, center of mass (members), and collective
heading. The agents move according to $p_i[k+1] = p_i[k] + \delta\,v_i[k]$,
where $v_i[k]$ is the scaled velocity of agent $i$
\begin{equation}
  \small
  v_i[k] = H_i\,\hat{\beta}_i[k] + \frac{1}{N_i}\sum_{j \in \mathcal{N}_{i}} \left[\hat{\gamma}_{ij}[k] \left((1-H_i) - \frac{p_{0}^{2}}{|p_{i}[k] - p_{j}[k]|^{2}} \right)\right] + \frac{H_i}{N_i}\sum_{j \in \mathcal{R}_{i} \cup \{i\}}  \hat{\theta}_j[k] \;\;.
  \label{eq:BoBSwarm}
\end{equation}
Here, $\hat{\beta}_i = (\cos\beta_i,\sin\beta_i)$ is the bearing vector
towards the goal, $\hat{\gamma}_{ij} = (\cos\gamma_{ij},\sin\gamma_{ij})$ the
azimuth vector of agent $i$ towards agent $j$, and $\hat{\theta}_j$ the
heading of agent $j$ as defined before. The heading consensus is achieved here
on a subset of agents $\mathcal{R}_i$ that are within a distance $p_0$ from
agent $i$, making the heading consensus localized. The binary heading
consensus weight $H_i$ determines whether the agent moves toward a goal ($H_i
= 1$), or purely positions itself with respect to the other members of the
swarm ($H_i = 0$). Leader-follower behavior is established by setting the
position of agent $i$ (leader) as the goal for the other agents in the swarm.
The scaling factor of $\hat{\gamma}_{ij}$ is more influenced by close spacing,
leading to strong repulsion and collision avoidance. When the distance between
two agents is the equilibrium distance $p_0$, the corresponding scaling factor
of $\hat{\gamma}_{ij}$ is zero.
The time constant $\delta$ is a parameter that depends on the update rate and
the overall speed scaling. In general, the formulation \eqref{eq:BoBSwarm}
results in lattice-like swarm arrangements, with different types of avoidance
strategies (see Fig. \ref{fig:BoBavoidance}).

\subsection{Robotic Platforms}
We tested the technology on two different platforms detailed below.

\subsubsection{eBot}
The eBot is a commercial small-size differential drive robot developed by
EdgeBotix\footnote{\url{http://www.edgebotix.com/}}, a spin-off company
developing educational robots based on research robots developed by the SUTD
MEC Laboratory\footnote{\url{http://people.sutd.edu.sg/~erikwilhelm/}}. It is
equipped with 6 ultrasonic or infrared range finders, inertial measurement
unit (IMU), two wheel encoders and two light sensors. The maximum velocity of
the platform is $20$ cm/s. We developed an extended K\'alm\'an filter
(included in the eBot's API) to localize the robot in real-time based on its
IMU and wheel encoders.
\begin{figure}[htbp]
  \centering
  \begin{subfigure}{0.49\columnwidth}
    \centering
    \includegraphics[width=0.6\columnwidth]{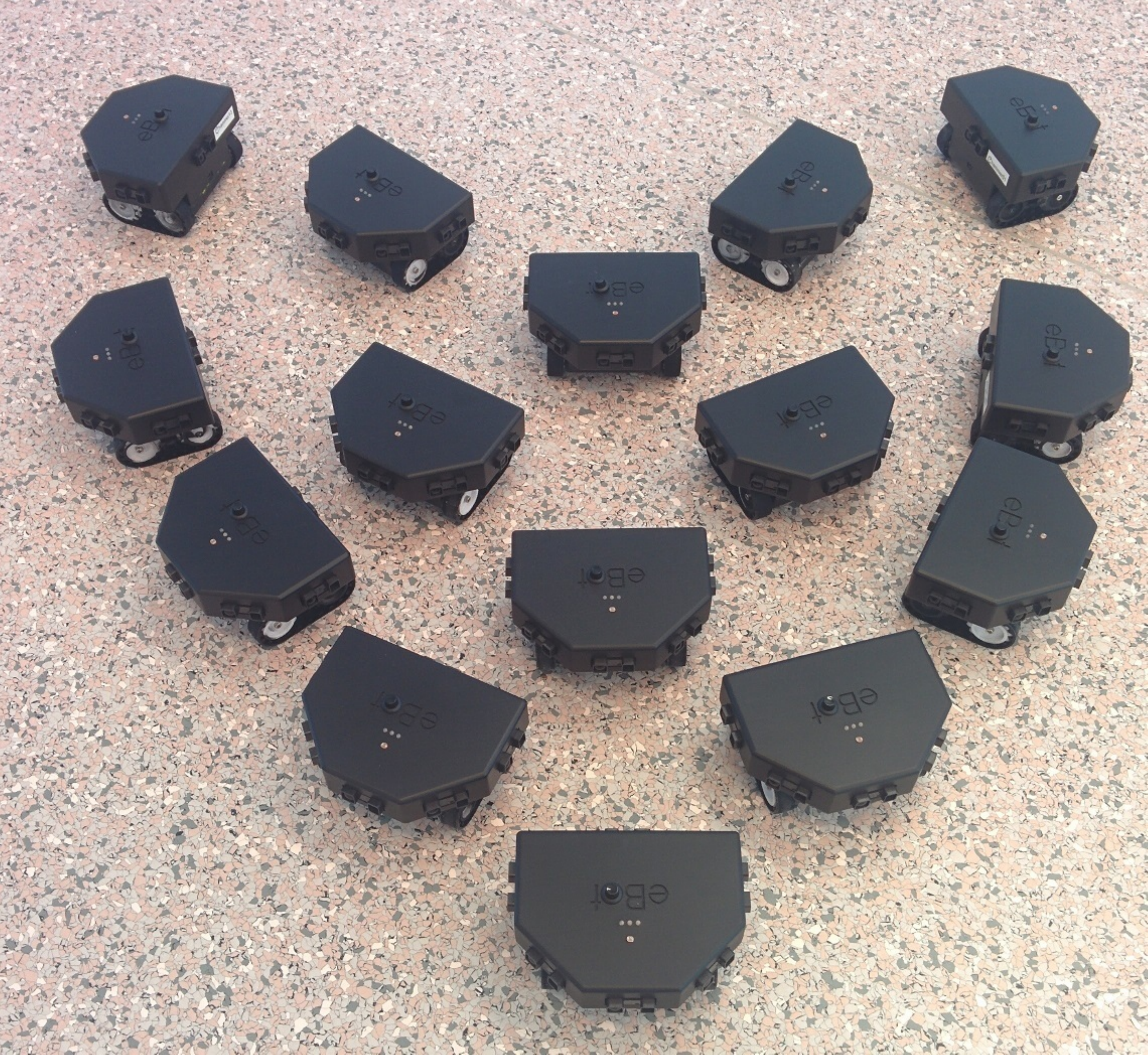}
  \end{subfigure}
  \begin{subfigure}{0.49\columnwidth}
    \centering
    \includegraphics[width=1.1\columnwidth]{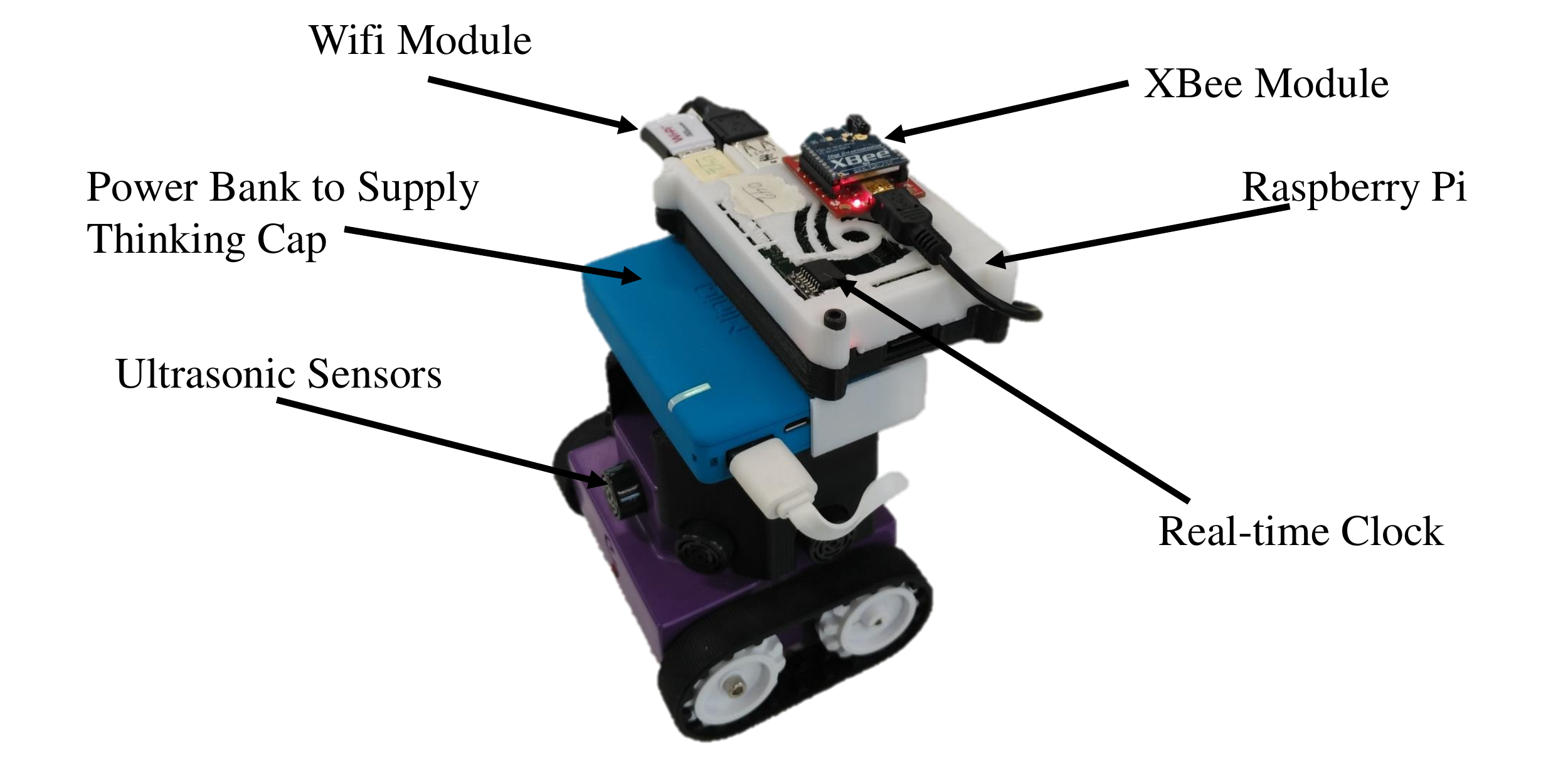}
  \end{subfigure}
  \caption{Right: the eBot with ultrasound sensors crowned with the
    swarm-enabling technology. A 3D printed structure is hosting the Raspberry
    Pi, and XBee module, along with a power bank serving as the power
    source. The WiFi module is used exclusively for monitoring and updating
    purposes, and is not required for the autonomous operation of the
    robot. Left: Swarm of $15$ eBots with IR sensors.}
  \label{fig:eBot platform}
\end{figure}
\subsubsection{Autonomous Surface Vehicle}
\begin{figure}[htbp]
  \centering
  \includegraphics[width=0.25 \columnwidth]{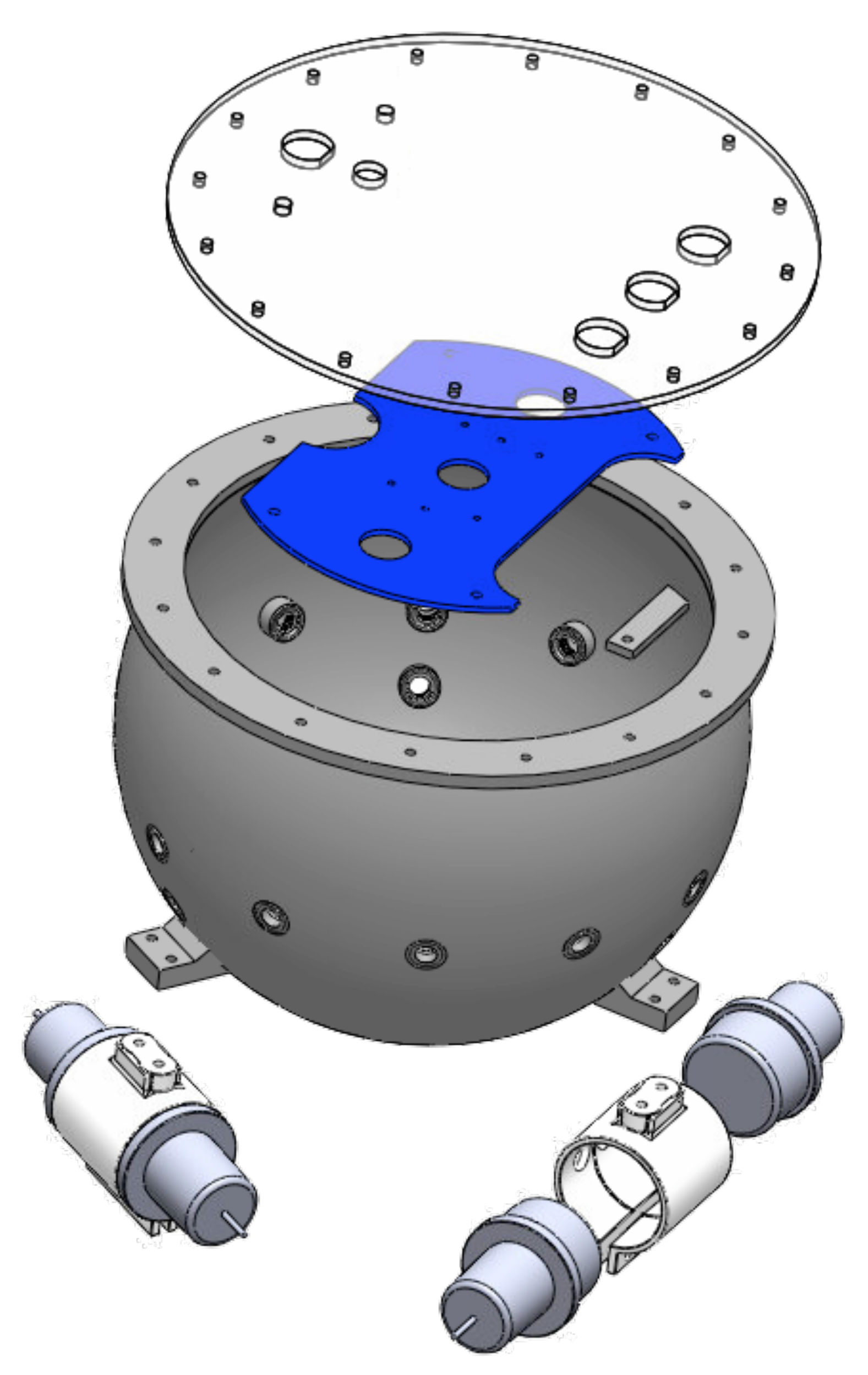}
  \includegraphics[width=0.35 \columnwidth]{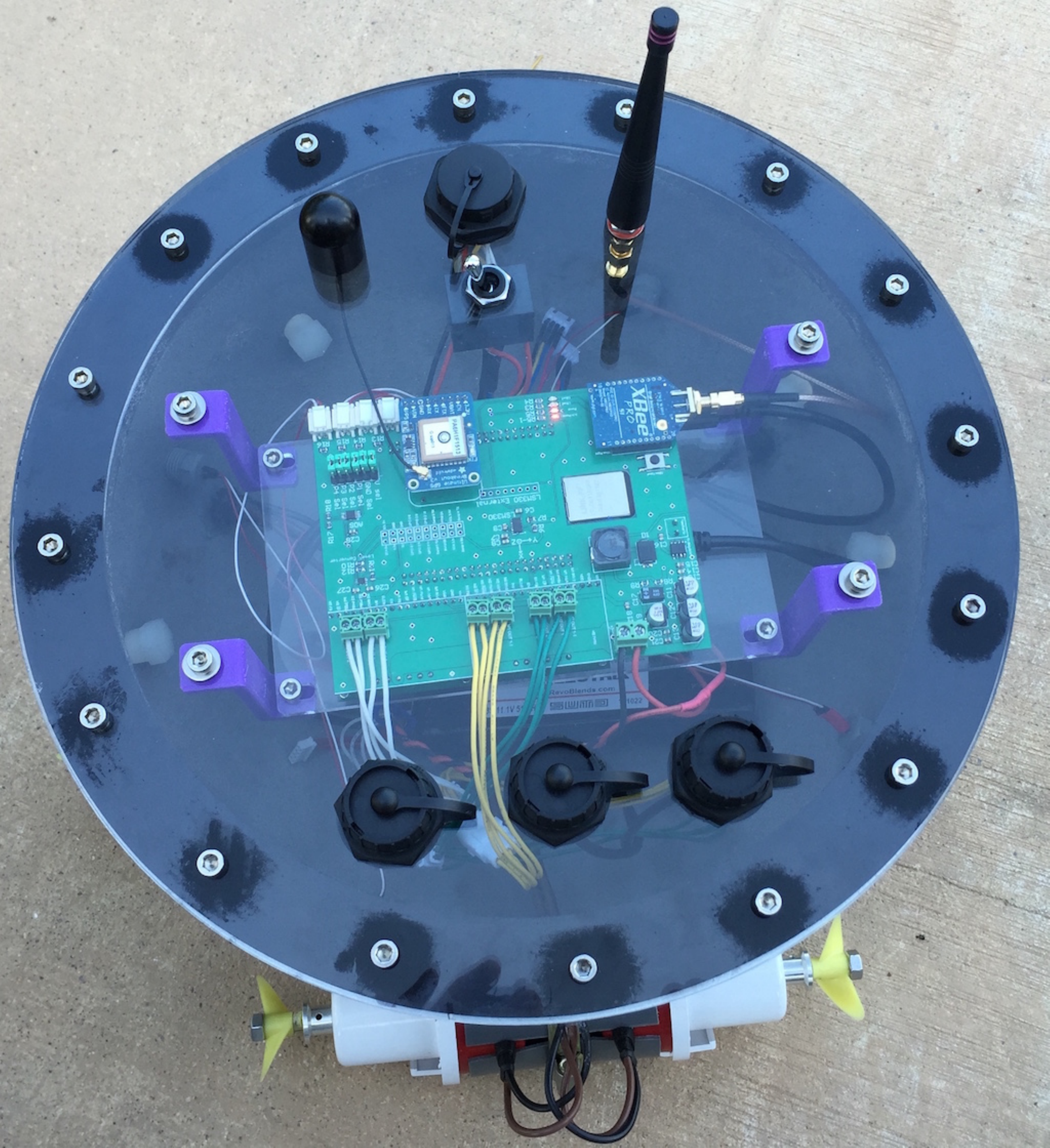}
  \caption{A cast aluminum hull mitigates magnetic interference with sensitive
    MEMS sensory elements used to reference buoy orientation.  Many standard
    expansion ports along the perimeter allow for the addition of
    environmental measuring devices seamlessly in the field.  A large access
    cover at the top provides easy access for maintenance as needed.  All
    components are housed securely within the watertight hull and can be seen
    on the top view (top right corner).}
  \label{fig:BoB_Rendering}
\end{figure}

The ``BoB'' (for ``Bunch of Buoys'') system is a distributed multi-robots
effort based on a small developmental surface craft
(Fig. \ref{fig:BoB_Rendering}) initially developed at MIT to perform
collective environmental sensing. (Videos of the collective sensing with more
than 50 units during a field test are available online~\footnote{Dynamic
  Environmental Monitoring using Swarming Mobile Sensing
  Buoys:~\url{https://youtu.be/Qe-wZOi3ONs}}~\footnote{51 Networked Buoys
  Swarming:~\url{https://youtu.be/fhg1rIX_y3A}}~\footnote{Dynamic Area
  Coverage (Geofencing) Field Test:~\url{https://youtu.be/hlBNjHS_Q7s}}.) It
is equipped with a Global Positioning Satellite (GPS) receiver, MEMS compass
and $3$-axis accelerometer.  Omni-directional design concepts lead to the
vectored propulsion system, allowing for maximum agility with near
instantaneous direction changes.  The maximum velocity of the platform is up
to $1$ m/s, although design considerations are more concerned with positioning
than transit.

\begin{figure}[htbp]
  \centering
  \includegraphics[width=0.6\columnwidth]{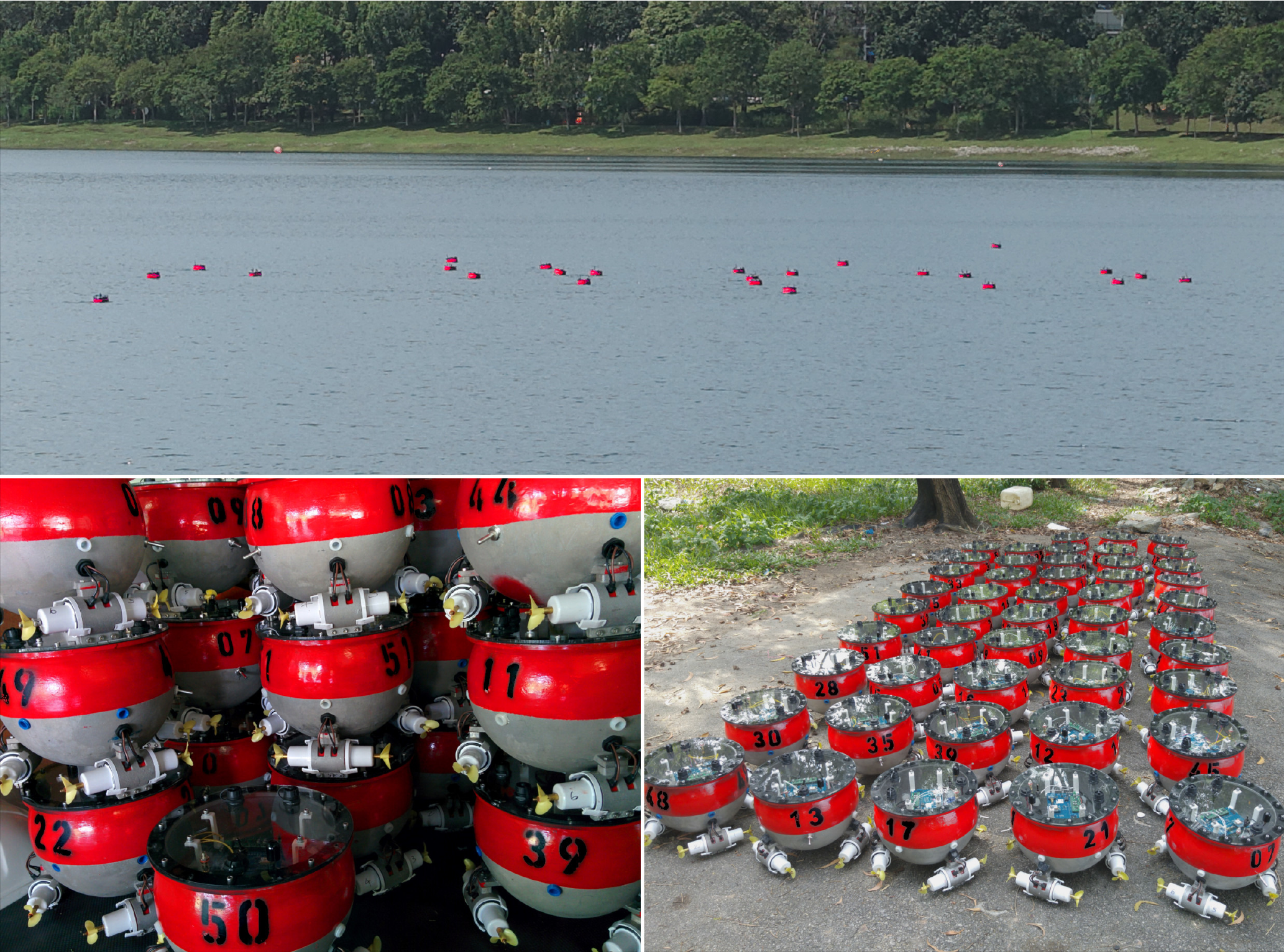}
  \caption{A small fleet of autonomous surface vehicles: (Top panel) 25 buoys
    collectively operating at Bedok Reservoir, Singapore; (Bottom left panel)
    Buoys stacked up during transportation to the field site; (Bottom right
    panel) 48 buoys lined up before deployment.}
  \label{fig:montage}
\end{figure}

Finally, the buoy may be equipped with multiple sensors in order to monitor
its environment. The fleet of buoys (BoB) (Fig.~\ref{fig:montage}) is not
constrained to homogeneity, and some units may be equipped with various
sensors in order to provide varying levels knowledge about the surrounding
waters.

\subsection{Software}

The connection between robot control, distributed communications management,
and collective behavior is established by means of a Python module called
\textit{marabunta}\footnote{See
  \url{http://www.github.com/david-mateo/marabunta}} that we designed with
modular development of swarming behaviors in mind and that emphasizes
portability and ease of deployment in different platforms.

With this module, the interaction with the robot is abstracted at a low-level
through a platform-dependent class (the ``body'' of the agent).  In the same
fashion, the management of communications is also handled by another low-level
and platform-dependent ``network'' class.  Both of these classes are
independent from one another and are also independent from the high-level
behavior of the swarm, defined as the ``behavior'' class.

Each different robot platform requires its own \texttt{body} class.  This is
designed following the assumption that the processor running the code has the
capacity to interact with the robot such that it can (\textit{i}) send
commands to move it in space, (\textit{ii}) request the data needed for
localization of the unit, and (\textit{iii}) access the information gathered
by the robot about its local environment typically by means of a sensor suite.
The swarm-enabling unit (SEU for short and shown in Fig.~\ref{fig:block
  diagram of the system}) does not require total access to the inner workings
of the robot or deep knowledge of its specifications.  For instance, if one is
using commercial robots such as the eBot or e-puck~\citep{puck}, establishing
a Bluetooth connection and using the provided API to send commands to the
robot is enough to use this technology to deploy a swarm of robots. The
\texttt{body} classes for both the eBot and the e-puck are included in \texttt{marabunta}.
%

%
The communication between agents is handled by a separate \texttt{network}
class.  Separating this from the control of the body allows to pair different
robot platforms with different communication protocols.  This implementation
assumes that the processor has the capacity to interact with a communication
module capable of (\textit{i}) broadcasting messages to nearby agents, and
(\textit{ii}) reading the messages broadcast by other agents.  There is no
need for the communication module to be able to handle directed communication
or recognition of nearby nodes.  The \texttt{marabunta} module provides a
network class for the Digimesh protocol, so that one can connect an XBee
module in serial to the computer running the software to get out-of-the-box
distributed communications between agents afforded with the present
technology.

The collective behavior of the swarm is a result of the agents' individual
behavior, implemented in a separate \texttt{behavior} class that is
hardware-agnostic.  Since the behavior is independent of the robot used, this
technology allows to have heterogeneous swarming where different robots
perform the collective behaviors described in the previous section.
Additionally, since the behavior is also independent of the communication
between agents, a swarm can have heterogeneous behaviors where different
agents follow different behaviors.  A friction-less platform to experiment
with heterogeneous swarming can yield interesting novel emergent behaviors
arising from the combination of behaviors (for example, combining some ratio
of agents performing consensus with others performing perimeter defense makes
the swarm split in sub-groups).

The modular design of the technology allows for fast experimentation.  One can
use the included \texttt{MockBody} and \texttt{MockNetwork} to design
collective behaviors iteratively using simulations.  From there, one can
change the body to the proper robot controller and do some preliminary tests
from a single, central control terminal (by keeping a ``mock network'' in the
terminal).  Finally, by affording each robot with a SEU with the desired
behavior and a proper communication module, the system is ready to perform
swarming experiments in a truly decentralized fashion.

\subsection{Integration}

In order to illustrate and exemplify the full integration of our SEU, we
detail in this subsection the actual implementation used for the search and
exploration experiments whose results are presented in Sec.~\ref{sec:sae}.

\begin{figure}
  \includegraphics[width=\linewidth]{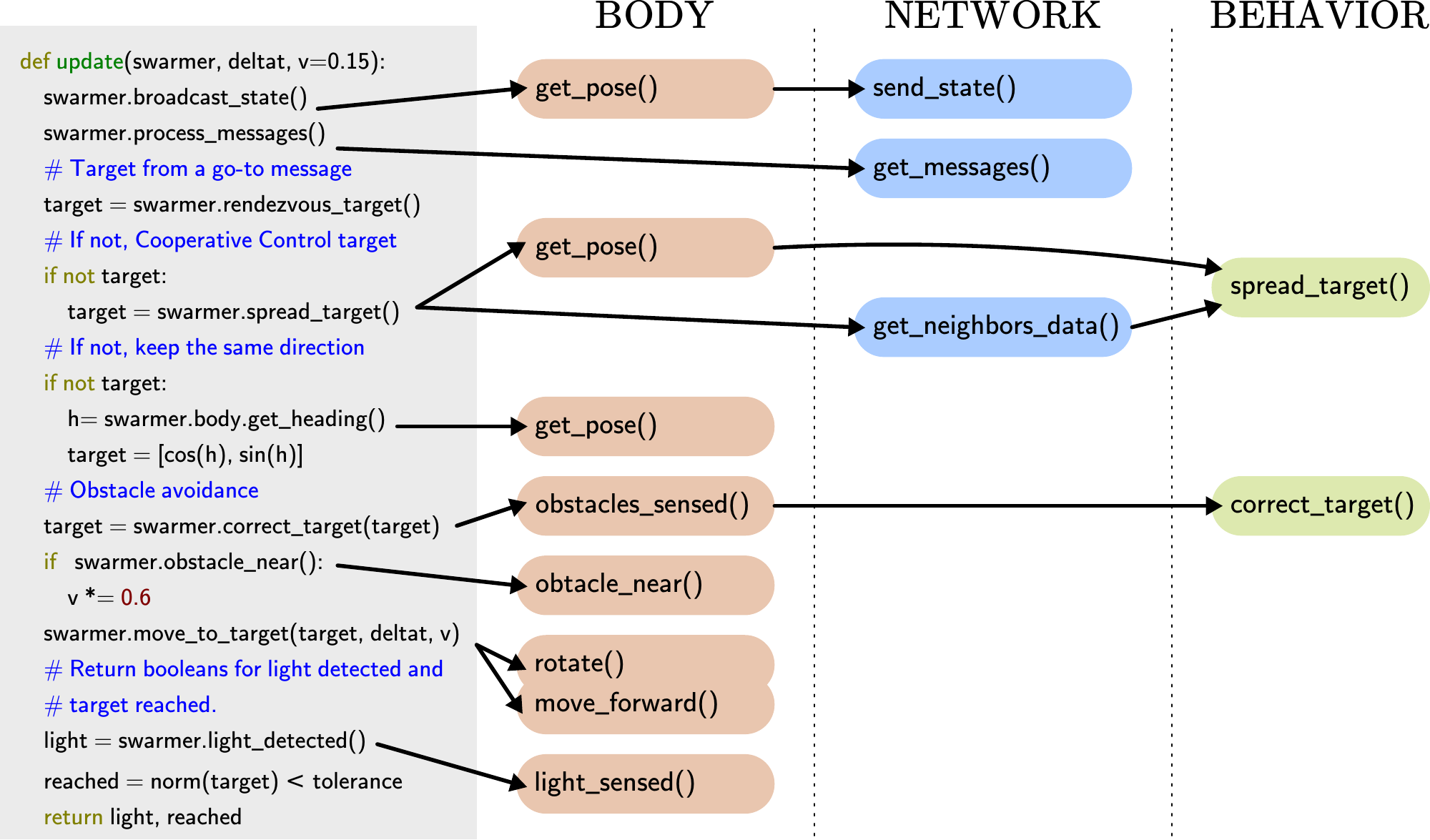}
  \caption{An example of the behavior of the agents defined with an update
    function that is called in a loop.  The update function calls different
    method of the body (red), network (blue), and behavior (green) of the
    agent.}\label{code_update}
\end{figure}

The natural starting point is implementing the cooperative control algorithm
described in Sec.~2.3.2 as a behavioral \texttt{behavior} class.  By
generating some artificial data for walls and position of the target to
locate, one can readily simulate the behavior of the swarm and iterate the
design of the algorithm by providing the behavior with a \texttt{MockBody} and
\texttt{network} classes.  For the search and exploration experiment, the
behavior of a swarming agent is defined by the update rule presented in
Fig.~\ref{code_update}.  This figure highlights the dependencies with the
methods implemented in the \texttt{body}, \texttt{network}, and
\texttt{behavior} classes.

After iterating the design of the behavior via simulation, we equip a fleet of
10 eBots with a ``swarming-enabling unit'' (SEU) consisting of a Raspberry Pi
with Python and the \texttt{marabunta} module loaded on it, an XBee module, a
Bluetooth module, and a power bank.  To operate these robots one has to
establish a Bluetooth connection and send simple commands (such as activate
wheels with a certain power or read the values of each sensor) through their
dedicated Python API.  Implementing a \texttt{body} class for the eBot is, in
practice, a matter of expanding the provided API to obtain a comprehensive
interface to interact with the robot and return the sensor data in a
convenient, hardware-agnostic format (e.g. giving the estimated coordinates of
the obstacles detected, as opposed to the raw sensor data).

For this kind of experiments, one does not need most of the features of the
communication module; the XBee can be set in transparent mode and interfaced
purely by writing the messages to be sent and reading the received ones
through a serial connection.  This makes the \texttt{network} class
implementation for an XBee module quite straightforward.  The main task of
this class is to translate the received messages and properly structure the
data so that the other elements of the module can access it.

By switching the \texttt{MockBody} for the \texttt{eBotBody} and the
\texttt{MockNetwork} by the \texttt{XbeeNetwork}, the code used for simulation
can run on the SEU and make the swarm to autonomously perform search and
exploration.

\section{Results}

\subsection{Perimeter Defense}

\begin{figure}[htbp]
  \centering
  \begin{subfigure}{0.49\columnwidth}
    \centering
    \includegraphics[width=1.1\columnwidth]{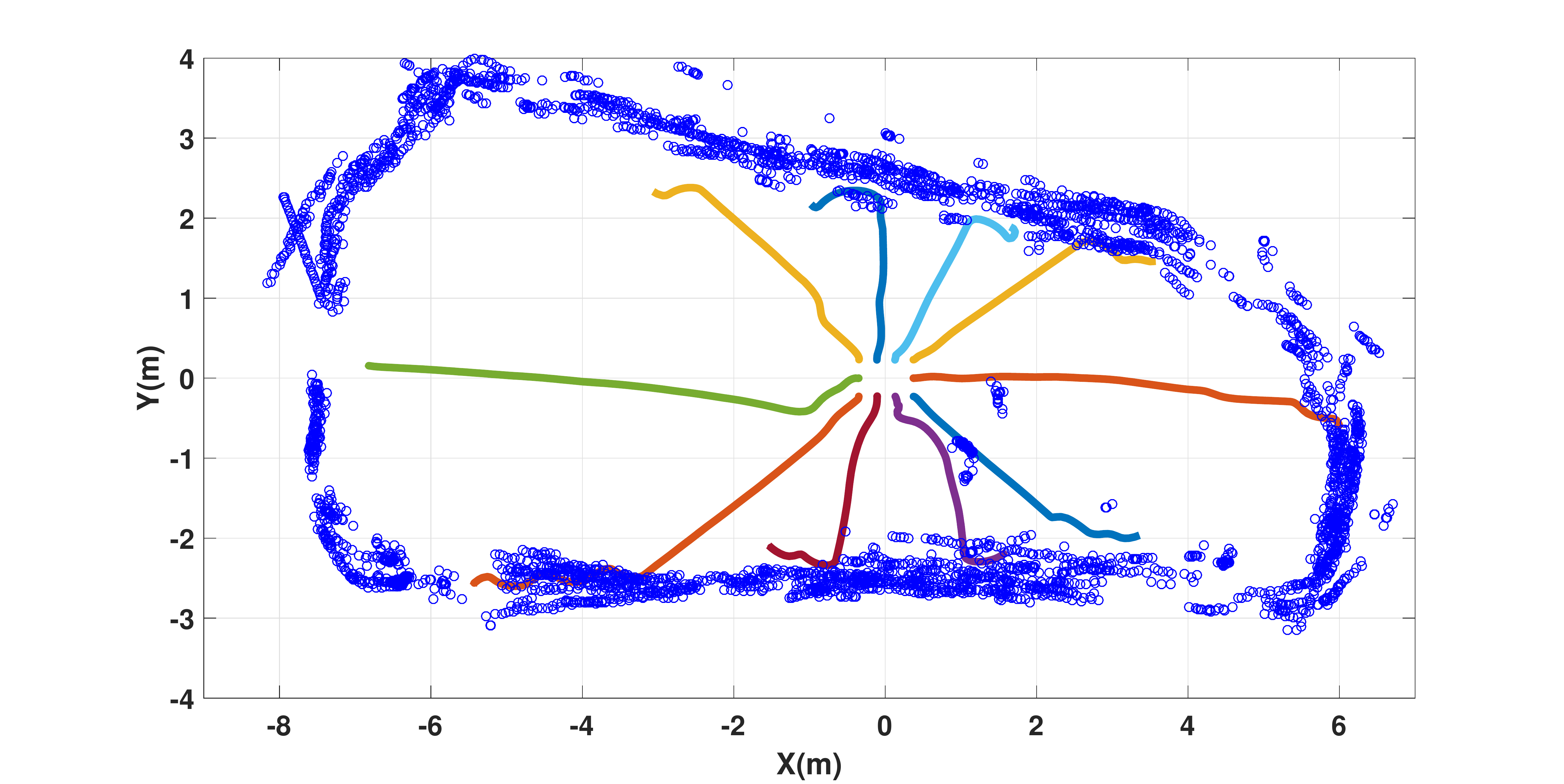}
    \caption{10 eBots}
  \end{subfigure}
  \begin{subfigure}{0.49\columnwidth}
    \centering
    \includegraphics[width=1.1\columnwidth]{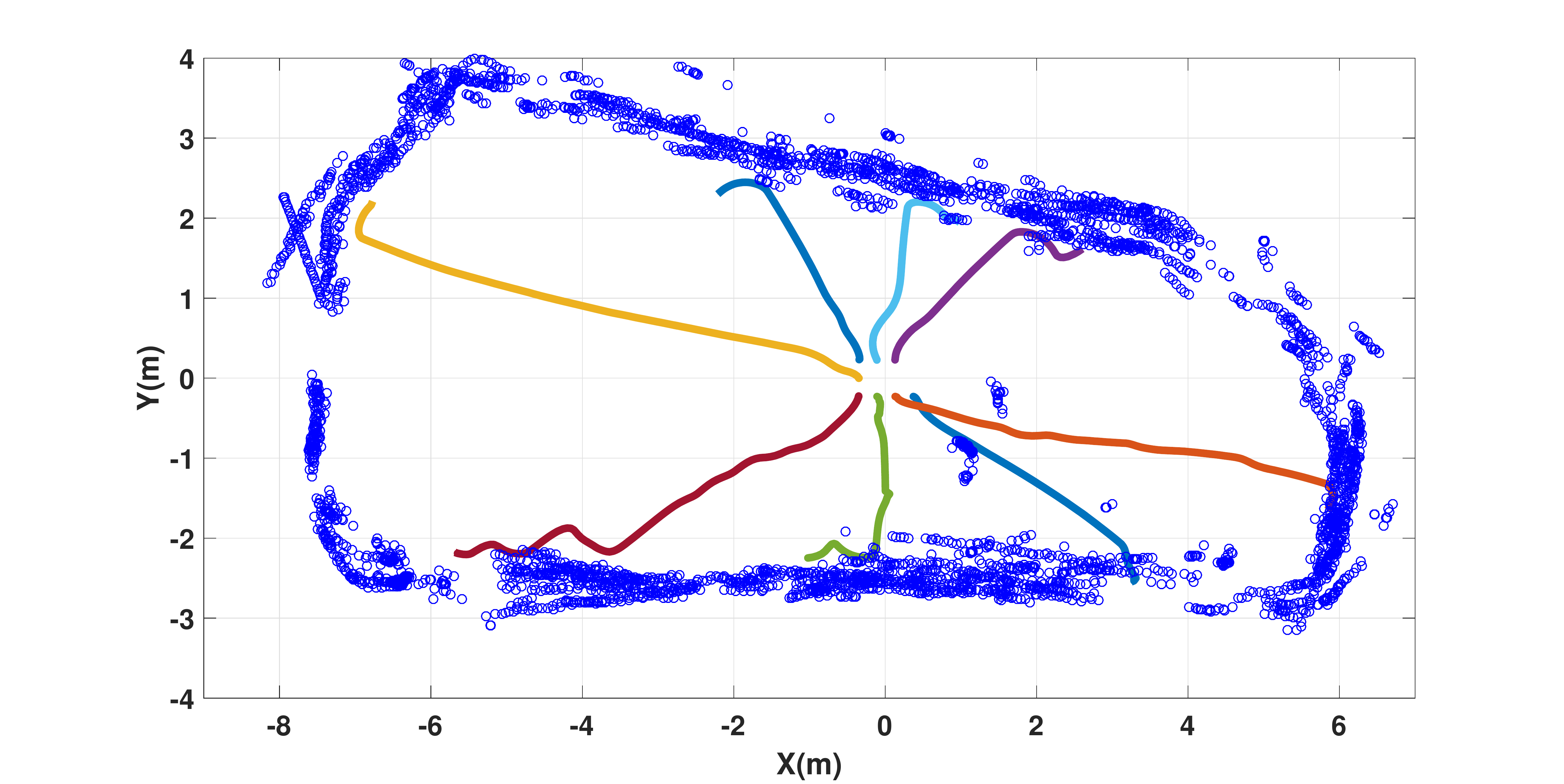}
    \caption{8 eBots}
  \end{subfigure}
  \\[\baselineskip]
  \begin{subfigure}{0.49\columnwidth}
    \centering
    \includegraphics[width=1.1\columnwidth]{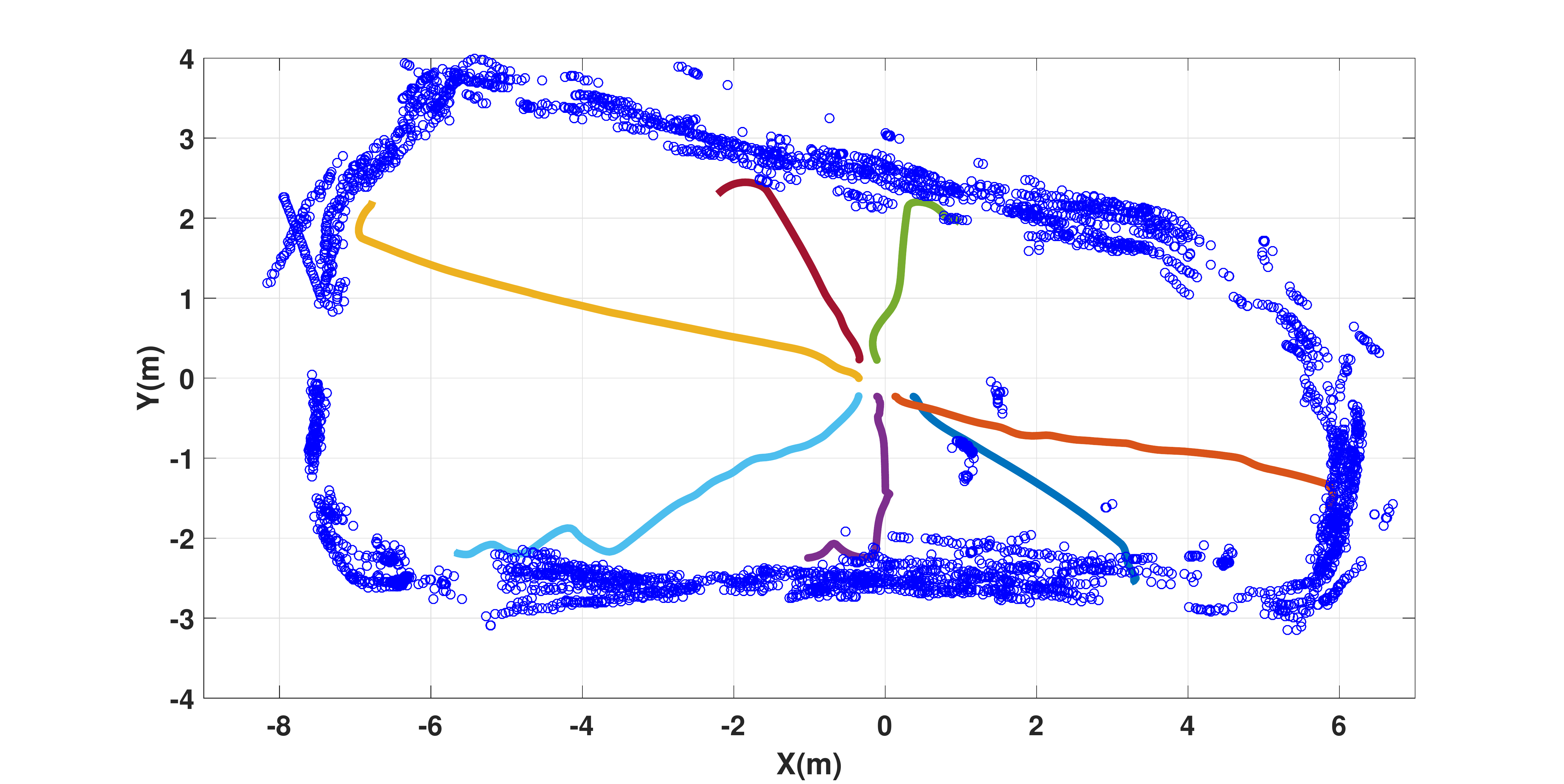}
    \caption{7 eBots}
  \end{subfigure}
  \begin{subfigure}{0.49\columnwidth}
    \centering
    \includegraphics[width=1.1\columnwidth]{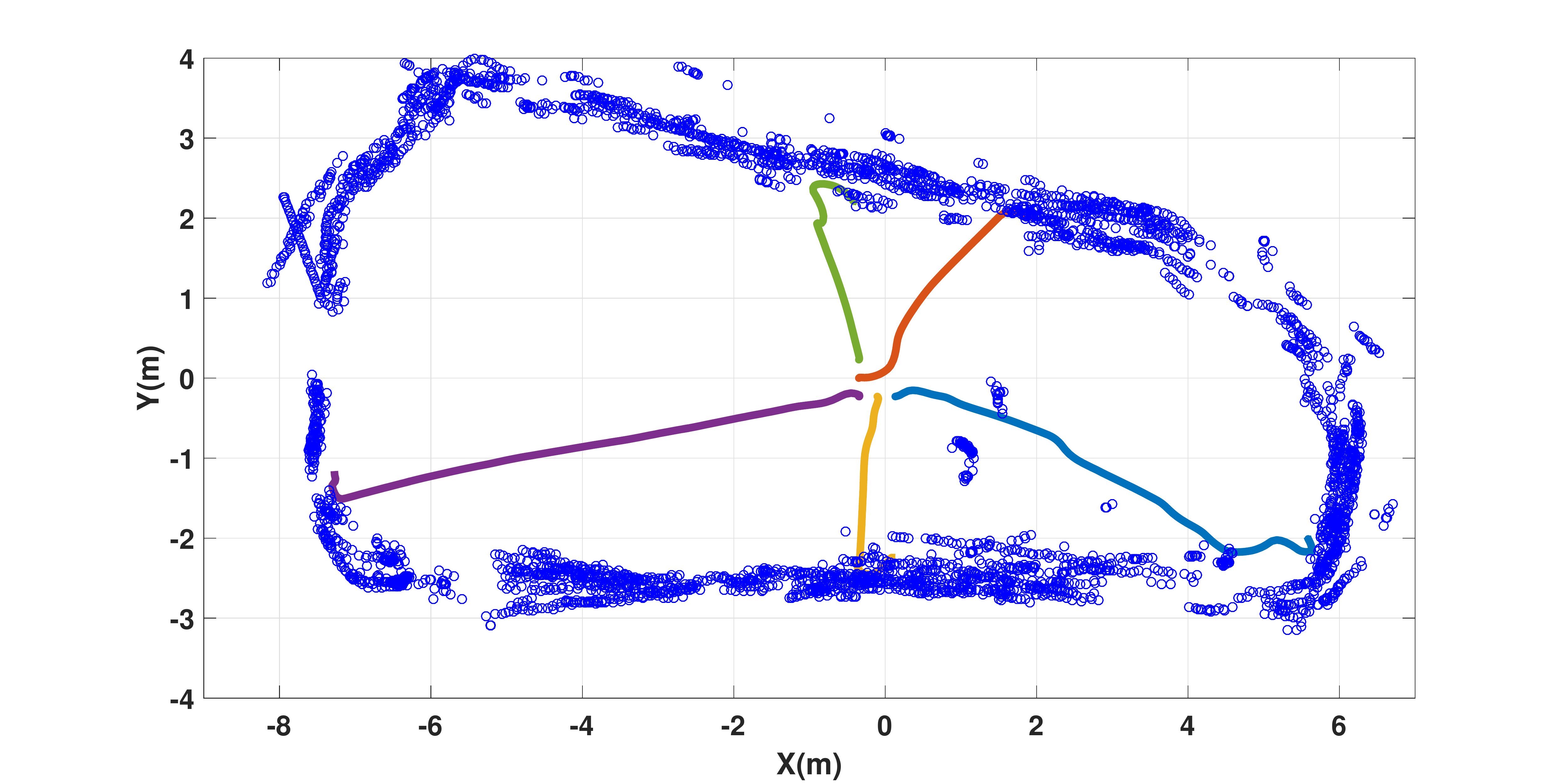}
    \caption{5 eBots}
  \end{subfigure}
  \caption{Perimeter defense algorithm using 5, 7, 8 and 10 eBots. The
    experiment is performed in a room with approximate dimension of
    $13.5\times 6.2$~m. All robots start with zero initial heading, and
    information about all initial positions and headings. Each robot uses an
    extended K\'alm\'an Filter to estimate its position and heading based on
    on-board sensors. This information is shared among neighbors leading to
    the decentralized computation of a target heading based
    on~\eqref{eq:perimeter defense}. }
  \label{fig:expansion}
\end{figure}

Experimental results regarding the perimeter defense algorithm discussed in
Section \ref{sec:perimeter defense} for different number of robots are shown
in Fig.~\ref{fig:expansion}. One can observe the dynamic decision making
feature of the swarm algorithm in Fig. \ref{fig:expansion}-(a) to
\ref{fig:expansion}-(d). Each individual decides on its target heading based
on the information received from other agents in the network. For this reason,
the target direction of robots are vastly different with different number of
robots. We also remark that one of the most important features of a swarm is
its scalability. This feature can be observed in Fig. \ref{fig:expansion} that
by reducing the number of robots from 10 to 5, the swarm is still capable of
covering the perimeter in the best possible manner. Robustness has been tested
with the same scenario with the forced removal of a number of units while
performing the experiment. In such a case, the whole swarm dynamically reacts
to this change and tries to cover the area removed robots were supposed to
cover.  As mentioned earlier, the Markovian nature of the swarming algorithms
enables the swarm to adapt instantaneously to any change in the environment or
configuration, including the addition or removal of agents and modifications
of the experimental arena.

\subsection{Search and Exploration}
\label{sec:sae}

\begin{figure}[htbp]
  \centering
  \includegraphics[width=0.7 \columnwidth]{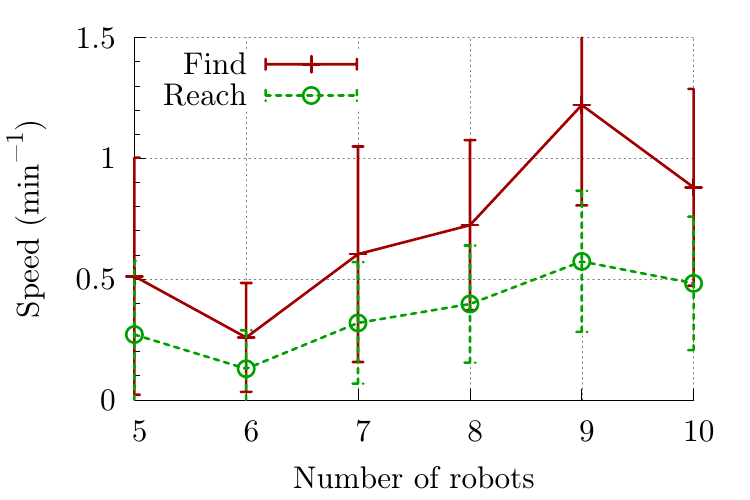}
  \caption{Speed at which one agent finds the target (solid red) and at which
    the swarm reaches the target (dashed green) as a function of the number of
    agents in the swarm.  Each data point is averaged over 10 runs. The error
    bars show the variance over these runs.  }
  \label{fig:perimeter speed}
\end{figure}

The second experiment is regarding a \emph{search and exploration} task in
which robots try to collectively find a target in the environment. The target
in our setup is a light source in an unknown environment. This experiment is
divided into two phases: \emph{(i)} the robots perform a perimeter defense
until one finds the light source and informs its neighbors about the location
of the target, and \emph{(ii)} the robots perform a \emph{rendezvous in space}
to collectively navigate to the location of the detected light source. The
algorithm was tested with up to 10 eBots. We ran ten experiments for each
number of agents to better quantify the effectiveness of the swarming
behavior. Fig.~\ref{fig:perimeter speed} shows the convergence speed---inverse
of the time it takes to find the target and navigate to its location---with
respect to the number of robots. Scalability of the swarm can be seen from the
upward trend of the graphs in Fig. \ref{fig:perimeter speed}: in most cases,
incorporating more agents in the experiment speeds up the search and
exploration task (watch video online~\footnote{Swarm Robotics: Collective
  Search \&
  Exploration:~\url{https://www.youtube.com/watch?v=JzbWV1sfZ-A}}). The same
experiment was expanded to allow for 14 robots swarming in 5 distinct rooms
(watch video online~\footnote{Simultaneous Collective Exploration of 5 rooms
  by a Swarm of 14 Terrestrial
  Robots:~\url{https://www.youtube.com/watch?v=0tsAx6TDy-Q}}).

\subsection{Heading Consensus}

\begin{figure}[htbp]
  \centering
  \includegraphics[width=0.8 \columnwidth]{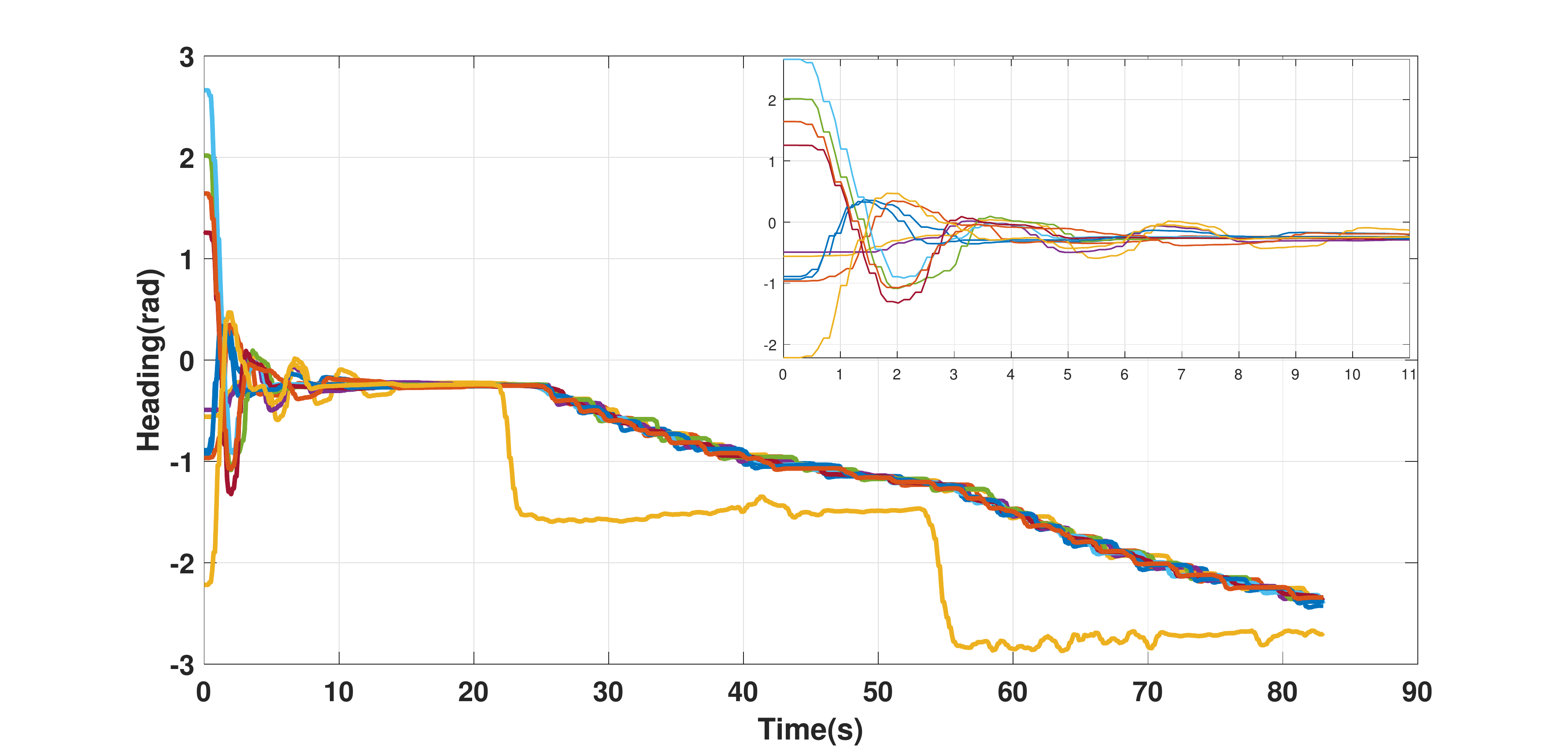}
  \caption{Heading consensus algorithm tested with a swarm of $10$ eBots. Each
    robot starts with a random heading. After a number of iterations, the
    swarm agrees on a common value for its heading. At time $t=22$~s, one of
    the robots stops participating to the consensus algorithm and we are in
    the presence of a leader-follower swarm, thereby forcing the swarm to
    converge to the leader's heading. The sub-figure is a zoomed-in view of
    the swarm heading from $t=0$ to $t=11$~s. }
  \label{fig:heading consensus}
\end{figure}

Figure~\ref{fig:heading consensus} presents the experimental results of the
heading consensus algorithm discussed in subsection \ref{sec:heading
  consensus} using $10$ agents.  This experiment consists of two parts: (1)
robots with random initial heading converge to a common heading, and (2) a
robot is forced not to follow the swarm heading. This can be interpreted as a
leader-follower configuration where the forced robot plays the role of the
leader in the swarm and forces the swarm to follow its heading.

\subsection{Aggregation and Leader-Follower} 

\begin{figure}[htbp]
  \centering
  \begin{subfigure}{0.51\columnwidth}
    \includegraphics[width=0.99 \columnwidth]{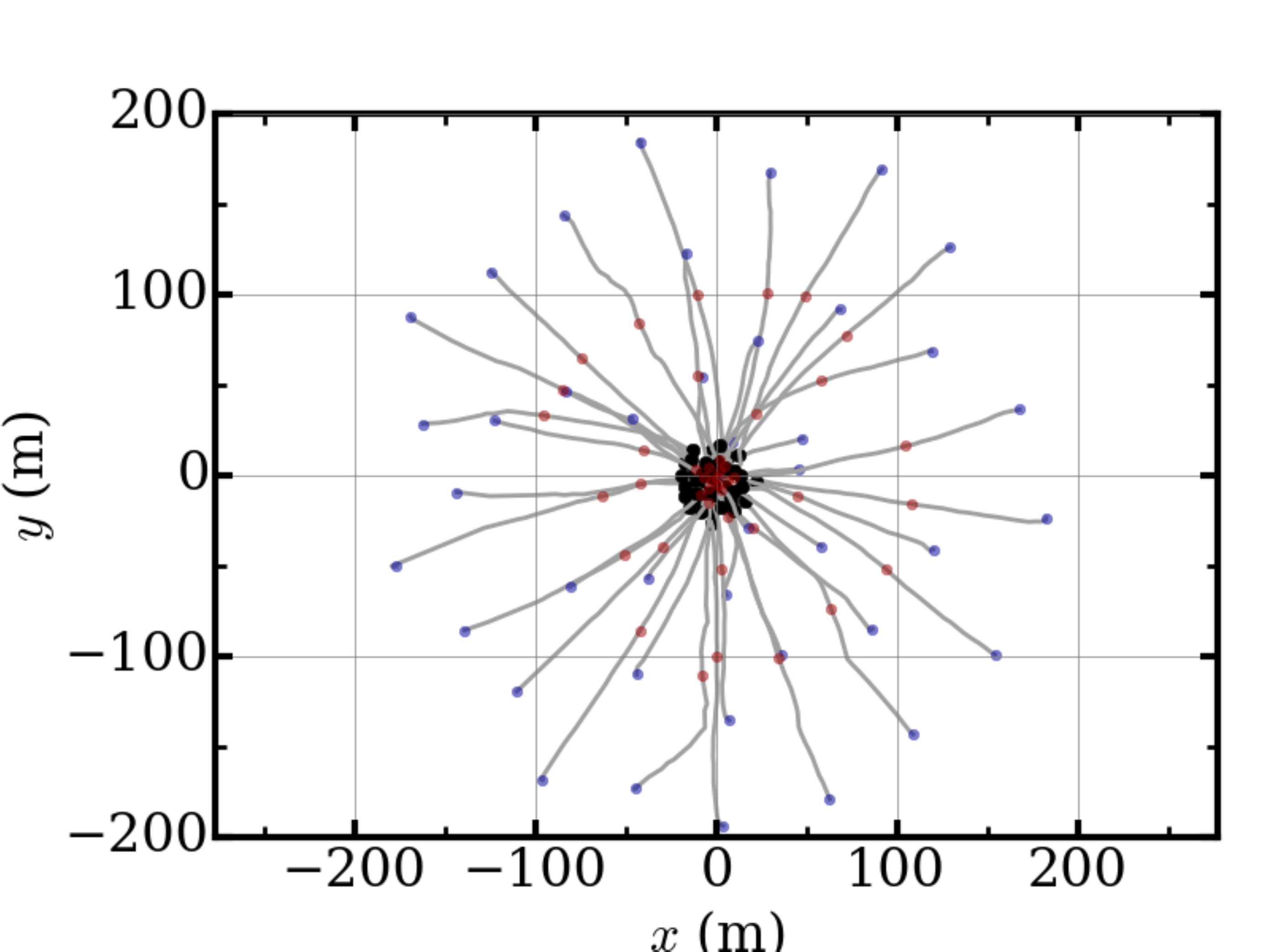}
  \end{subfigure}
  \begin{subfigure}{0.46\columnwidth}
    \includegraphics[width=0.99 \columnwidth]{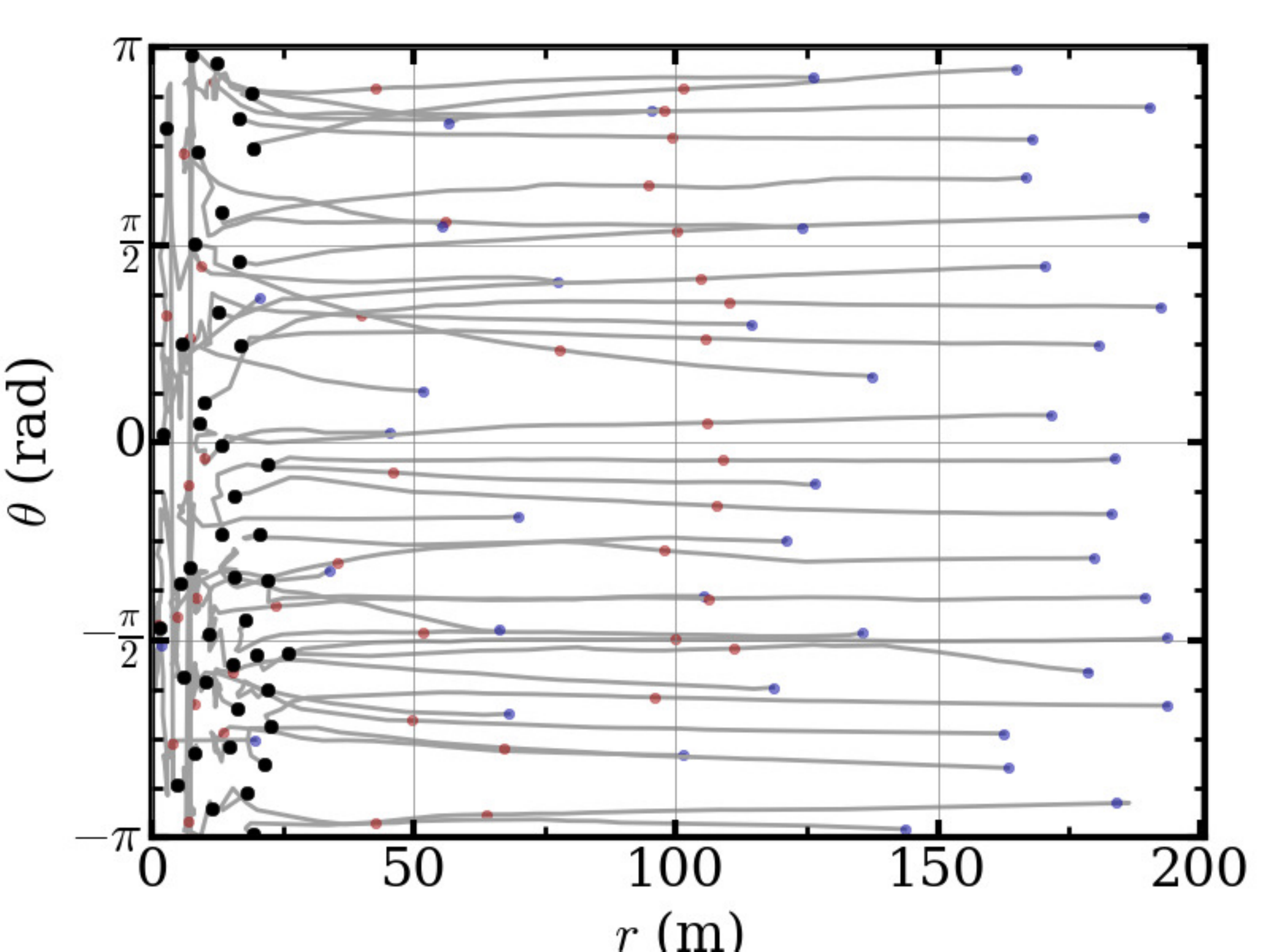}
  \end{subfigure}
  \caption{Field tests of aggregation behavior of a swarm of 45 buoys,
    conducted on a $\sim 1$ km$^2$ body of water on a moderately windy day.
    Left: Buoy trajectories for aggregation behavior. The
    aggregation test consisted of a single aggregation event ($p_0$ from $50$
    m to $5$ m), where the relative bearing to the group center is
    maintained. The entire aggregation event takes just over 8 minutes, in
    which time the coverage area is decreased from $\approx125,000$~m$^{2}$ to
    to $1,250$~m$^{2}$. Right: Buoy heading $\hat{\theta}_i$ as a function of
    distance $\hat{r}_i$ from the group center for the same aggregation
    event. Blue dots indicate the initiation position of the buoys, while red
    and black indicate the 50\% and terminal positions respectively.
    \label{fig:BoB1}}
\end{figure}

\begin{figure}[htbp]
  \centering
  \includegraphics[width=0.55 \columnwidth]{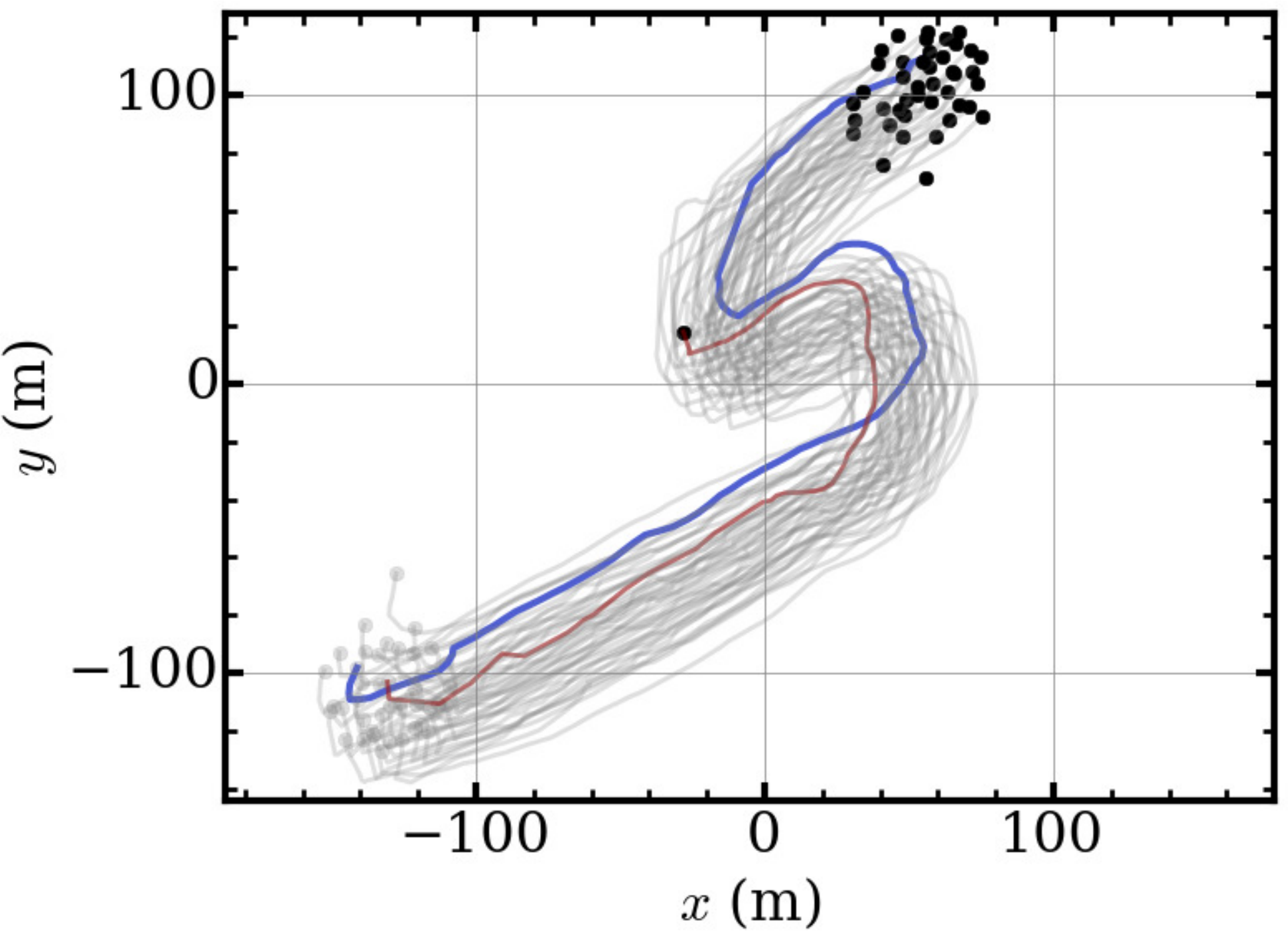}
  \caption{Field tests of leader-follower behavior of a swarm of 45 buoys with $p_0 = 5$~m.
   The group traverses $\sim 400$\,m in calm seas. The trajectory of the
   leader buoy is marked with a blue trail and that of the followers with a
   gray trail. The red trail highlights the trajectory of a buoy lagging
   behind due to low battery. Even though
   there are a number of degraded members, the system successfully executes the intended
   behavioral exercise.  
   \label{fig:BoB2}}
\end{figure}

Figure~\ref{fig:BoB1} shows some preliminary results of the aggregation
behavior and Fig.~\ref{fig:BoB2} of a leader-follower behavior according to
Eq.~\eqref{eq:BoBSwarm} for a swarm of $N=45$ buoys. For the aggregation test
($H_i \equiv 0$, $i = 1,\ldots,N$), the buoys are initially self-assembled in
a loose arrangement, and then the equilibrium distance $p_0$ is reduced; the
spreading follows by enlarging $p_0$. For the leader-follower behavior ($H_i
\equiv 1$, $i = 1,\ldots,N$), one buoy is driven along a given path, and its
position is given as the goal for all other buoys in the swarm. The
experiments show that lattice-like arrangement is persistent. They further
confirm that the mesh network strategy scales robustly for a swarm of this
size (Videos of these experiments are available online~\footnote{Dynamic
  Environmental Monitoring using Swarming Mobile Sensing
  Buoys:~\url{https://youtu.be/Qe-wZOi3ONs}}~\footnote{51 Networked Buoys
  Swarming:~\url{https://youtu.be/fhg1rIX_y3A}}~\footnote{Dynamic Area
  Coverage (Geofencing) Field Test:~\url{https://youtu.be/hlBNjHS_Q7s}}).

\section{Conclusions}

In this report, we presented the design of integrated hardware and software
tools enabling a wide range of multi-robot systems to collectively operate in
a fully distributed manner. Although our proposed swarm-enabling technology
requires some work to develop a basic interfacing with the selected mobile
platform, it nonetheless offers a general framework to assemble full-fledged
swarms from virtually any set of mobile robots. As already mentioned and shown
in the particular case of the e-puck, this basic interface can be extremely
simple and straightforward to put in place. In particular, this technology
adds decentralization features to the robotic platforms which is a key element
in a swarm and distributed robotics systems. Modular design of the software
library allows a swift transfer of the hardware toolkit onto vastly different
platforms. Moreover, the overall modular design of this swarm-enabling
technology facilitates possible future upgrade and evolution depending on the
particular requirements of any given swarm experiment, e.g. more powerful
computing unit to run swarming behaviors requiring on-the-fly machine
learning.  We reported experimental results regarding various swarm algorithms
on land and water surface platforms equipped with this swarm-enabling
technology.

Future research directions are towards including further swarm algorithms in
the software library and also the possibility to simulate newly designed
collective behaviors prior to their implementation onto the platforms. This
feature greatly facilitates the design and testing of new swarming behaviors.
Lastly, given the fact that our swarm-enabling technology can seamlessly
function with a host of different mobile robots, it should therefore
facilitate studies of heterogeneous swarming.

\section*{Disclosure/Conflict-of-Interest Statement}

The authors declare that the research was conducted in the absence of any
commercial or financial relationships that could be construed as a potential
conflict of interest.

\section*{Author Contributions}

R.B. and E.W. initiated the study and the development of the technology. All
authors designed the technology. M.C., D.M., G.T. and B.Z. implemented the
technology and carried out the field tests. D.M. developed the software
pack. All authors analyzed the results, wrote the main text and reviewed the
manuscript.

\section*{Acknowledgments}
This work was supported by grants from the Temasek Lab (TL@SUTD) under a Seed
grant \#IGDS S15 01021, from the Singapore Ministry of Education (MOE Tier 1)
grant \#SUTDT12015003, and by the National Research Foundation Singapore
under its Campus for Research Excellence and Technological Enterprise
programme. The Center for Environmental Sensing and Modeling is an
interdisciplinary research group of the Singapore MIT Alliance for Research
and Technology.


\end{document}